\documentclass[11pt]{article}

\usepackage[preprint]{acl}

\usepackage{times}
\usepackage{latexsym}
\usepackage{microtype}

\usepackage[UTF8,scheme=plain,fontset=fandol]{ctex}
\usepackage{subcaption}
\usepackage{amsmath,amssymb}
\usepackage{booktabs}
\usepackage{tabularx}
\usepackage{array}
\usepackage{graphicx}
\usepackage{placeins}
\usepackage[table]{xcolor}
\usepackage{multirow}
\usepackage[skins,breakable]{tcolorbox}
\usepackage[table]{xcolor}
\usepackage{enumitem}

\definecolor{profileblue}{RGB}{234,246,252}

\newcommand{\modelicon}[1]{%
  \makebox[1.15em][c]{%
    \raisebox{-0.18\height}{%
      \includegraphics[width=1.05em,height=1.05em,keepaspectratio]{fig/provider_icons/#1.png}}}%
  \nobreak\hspace{0.25em}}

\definecolor{promptnavy}{RGB}{35,55,78}
\definecolor{promptborder}{RGB}{196,212,226}
\definecolor{prompttagbg}{RGB}{235,243,249}
\newtcolorbox{ahaprompt}[2][]{%
  enhanced,
  colback=white,colframe=promptborder,
  colbacktitle=promptnavy,coltitle=white,
  fonttitle=\small\bfseries,fontupper=\small,fontlower=\small,
  title={#2},boxrule=0.5pt,arc=2mm,
  left=6pt,right=6pt,top=8pt,bottom=6pt,
  before skip=12pt,after skip=8pt,
  attach boxed title to top left={xshift=7pt,yshift=-2mm},
  boxed title style={colback=promptnavy,colframe=promptnavy,
    boxrule=0pt,arc=1.2mm,left=5pt,right=5pt,top=3pt,bottom=3pt},
  segmentation style={draw=promptborder,dashed,line width=0.5pt},
  before upper={\setlength{\parindent}{0pt}\raggedright},
  before lower={\setlength{\parindent}{0pt}\raggedright},
  #1}
\newcommand{\prompttag}[1]{%
  \tcbox[on line,box align=base,colback=prompttagbg,
    colframe=promptborder,boxrule=0.3pt,arc=2mm,
    left=4pt,right=4pt,top=0.7pt,bottom=0.7pt,boxsep=0pt]
    {\footnotesize\bfseries #1}}
\newcommand{\promptheading}[1]{%
  {\color{promptnavy}\footnotesize\bfseries #1}\par\smallskip}


\title{\: AhaPro: A Process-level Rebus benchmark for probing Aha moments in LLM reasoning}

\author{%
  Wei Guo\textsuperscript{1}\quad
  Lang Qin\textsuperscript{1}\quad
  Yundi Gai\textsuperscript{1}\quad
  Qinghua Xu\textsuperscript{1}\quad
  Zhi Wu\textsuperscript{2}\\
  \bfseries Wenjian Ding\textsuperscript{3,1,*}\quad
  Huijia Li\textsuperscript{1,*}\quad
  Zhenglu Yang\textsuperscript{1}\quad
  YAO ZHANG\textsuperscript{1,*}\\[0.6em]
  {\normalfont\small\textsuperscript{1}Nankai University}\\
  {\normalfont\small\textsuperscript{2}China Automotive Technology and Research Center}\\
  {\normalfont\small\textsuperscript{3}Juntendo University}\\[0.4em]
  {\normalfont\small\textsuperscript{*}Corresponding authors.}
}
\begin{document}
\addtocontents{toc}{\protect\setcounter{tocdepth}{-1}}

\maketitle
\raggedbottom


\begin{abstract}
Existing research on Aha moments in large language models has largely focused on reflective expressions or final revision outcomes, leaving their role in the visible reasoning process insufficiently characterized. Rebus puzzles encode answers through character forms, spatial arrangements, and homophony. Reinterpreting key details may trigger Aha moments and shifts in the reasoning process and candidate answers. We introduce AhaPro, a process-level benchmark that uses rebus puzzles to examine how models reorganize evidence and revise answers through Aha moments. The framework represents CoT as sequences of candidate answers, evidence, and verification states, using hierarchical metrics to assess conflict detection, state improvement, and subsequent adoption. Across eleven models, we observe frequent failures to adopt better states discovered during Aha moments, alongside adopted revisions that leave state quality unchanged or reduce it. Models also detect conflicts without subsequently achieving an effective revision. Further paired interventions suggest that prior reasoning context may hinder effective revision after conflict detection, while multi-branch search highlights answer selection as a bottleneck between correct-answer discovery and successful correction.
\end{abstract}


\section{Introduction}
\label{sec:introduction}

\begin{figure}[t]
    \centering
    \includegraphics[width=\columnwidth]{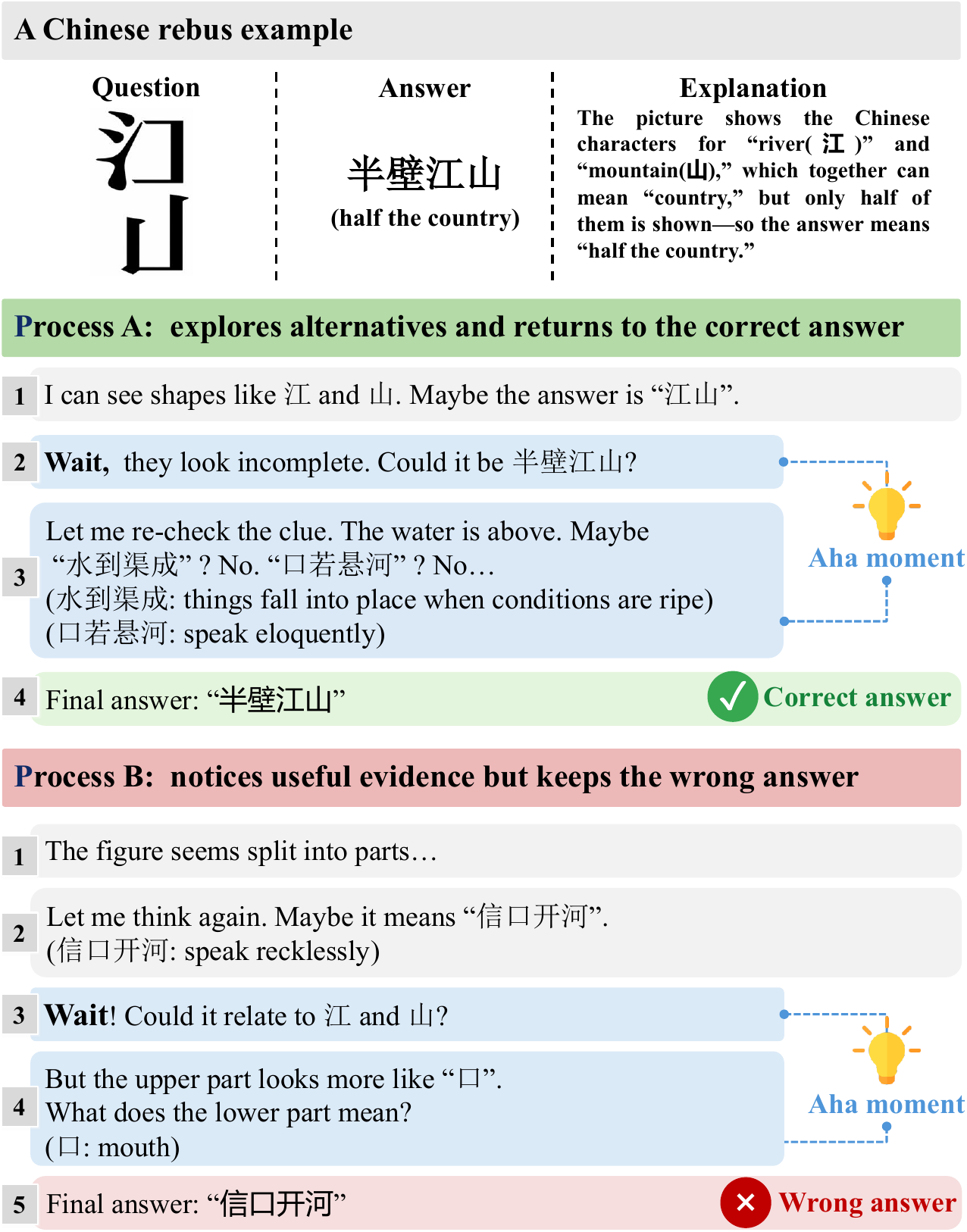}
    \caption{Illustrative CoT traces showing Aha moments without further improvement (A) and a better reasoning state that is not retained (B). These examples highlight differences between intermediate reasoning process, evidence adoption, and final-answer correctness.}
    \label{fig:rebus-aha-intro}
\end{figure}

Large language models sometimes use reflective expressions such as ``wait'' and ``reconsider'' in their reasoning traces and attempt to adjust their reasoning process. Such shifts are commonly referred to as Aha moments. Recent studies have found that Aha moments do not always improve final answer accuracy, and that some steps that appear to verify or revise an answer may have only a limited effect on the final answer \citep{daliberti2026illusion,zhao2025fakeaha}. Further investigation is needed to understand how these moments affect the reasoning process and whether they help models make effective corrections.

Existing studies typically identify Aha moments through reflective expressions or changes in the reasoning process, and assess their effectiveness through final answer accuracy \citep{huang2024selfcorrect,daliberti2026illusion,zhao2025fakeaha}. These outcome-based evaluations leave unclear how Aha moments affect intermediate reasoning states. An Aha moment may involve verbal reconsideration, conflict detection, evidence reorganization, or the formation of new candidate answers. Figure~\ref{fig:rebus-aha-intro} illustrates two outcomes: A explores alternatives and returns to the correct answer; B abandons relevant character clues and retains an incorrect answer. Even when a model discovers a better reasoning state, it may not adopt that state in subsequent reasoning. Evaluating these moments therefore requires tracing state changes, subsequent adoption, and final answer outcomes.

We select rebus puzzles because solving them requires integrating visual clues with linguistic knowledge and adjusting candidate answers during the reasoning process. In our cross-task comparison, Rebus averages 0.53 Aha moments per problem, compared with 0.32 for AIME2024 and 0.03 for MATH500. These results support the use of rebus puzzles for process-level analysis. Existing benchmarks, including REBUS \citep{gritsevskiy2024rebus}, COLUMBUS \citep{kraaijveld2025columbus}, Puzzled by Puzzles \citep{lee2025puzzled}, and the Chinese Pun Rebus Art Dataset \citep{zhang2025chinese}, primarily evaluate final answers or reasoning plausibility. They provide limited characterization of how candidate answers, evidence, and verification states evolve during reasoning.

We therefore introduce AhaPro, a process-level evaluation framework for analyzing Aha moments in large language model reasoning. It represents CoT as a sequence of reasoning states comprising candidate answers, evidence, and verification states, and tracks exploration paths and subsequent adoption. The framework distinguishes Aha candidates from substantive Aha events and classifies events according to changes in state quality and adoption. We further introduce a hierarchical metric suite to quantify conflict detection, better-state discovery, and subsequent adoption, and relate these processes to final answer outcomes. To support this evaluation, we curate 400 manually screened rebus puzzles, annotated with gold answers, evidence, and interpretation mechanisms.

Using this framework, we identify two key revision bottlenecks across eleven models. Specifically, 7.7\%-60.2\% of discovered better states are not adopted, while 32.7\%-66.9\% of traces with explicitly detected conflicts show no subsequent effective revision. Exploratory paired interventions further show that removing prior reasoning context significantly reduces incorrect-candidate retention, with only modest observed gains in answer repair rates. Multi-branch search also reveals selection failures even after the correct answer has been generated.

Our main contributions are as follows:

\begin{itemize}[leftmargin=*]
  \setlength{\itemsep}{-2pt}
\item We construct a dataset of Chinese and English rebus puzzles and provide gold answers, evidence, and interpretation mechanisms.

    \item We propose a process-level evaluation framework and metrics to characterize changes in the reasoning process, the discovery of better states, and their subsequent adoption during Aha moments.

    \item We characterize models' failures to revise effectively after detecting conflicts and their ``discovered but not adopted'' behavior. Paired interventions examine how prior reasoning context and multi-branch search affect observable revision behavior.
\end{itemize}

\section{Related Work}
\label{sec:related-work}

\paragraph{Rebus benchmarks.}
Rebus puzzles require models to combine visual clues, textual structure, spatial relations, and linguistic associations.
REBUS, COLUMBUS, Re-Bus, RebusBench, and Puzzled by Puzzles examine rebus solving across puzzle types and reasoning strategies \cite{gritsevskiy2024rebus,kraaijveld2025columbus,das2025rebus,kasaei2026rebusbench,lee2025puzzled}.
The Chinese Pun Rebus Art Dataset emphasizes homophony and cultural associations \cite{zhang2025chinese}; Reasoning Riddles distinguishes answer correctness from explanation quality \cite{movva2025reasoning}.
These evaluations focus primarily on final answers or completed explanations.
AhaPro adds process-level diagnosis by separating candidate reconsideration from substantive Aha and tracking A/E/V states and the discovery--adoption gap.

\paragraph{Aha Moments and CoT.}
CoT prompting elicits observable intermediate reasoning, including hypothesis formation, verification, rejection, and revision \cite{wei2022cot,kojima2022zeroshot}.
Visible CoT traces expose reconsideration and changes in reasoning strategy, often discussed as Aha moments \cite{deepseek2025r1}.
Such shifts do not consistently improve task performance, and apparently reflective steps may have limited functional effects \cite{daliberti2026illusion,zhao2025fakeaha}.
AhaPro evaluates these changes through observable state transitions and subsequent adoption.
We analyze CoT as behavioral outputs without assuming faithfulness to the model's internal computation \cite{turpin2023unfaithful}.

\begin{figure*}[!t]
    \centering
    \includegraphics[width=\textwidth]{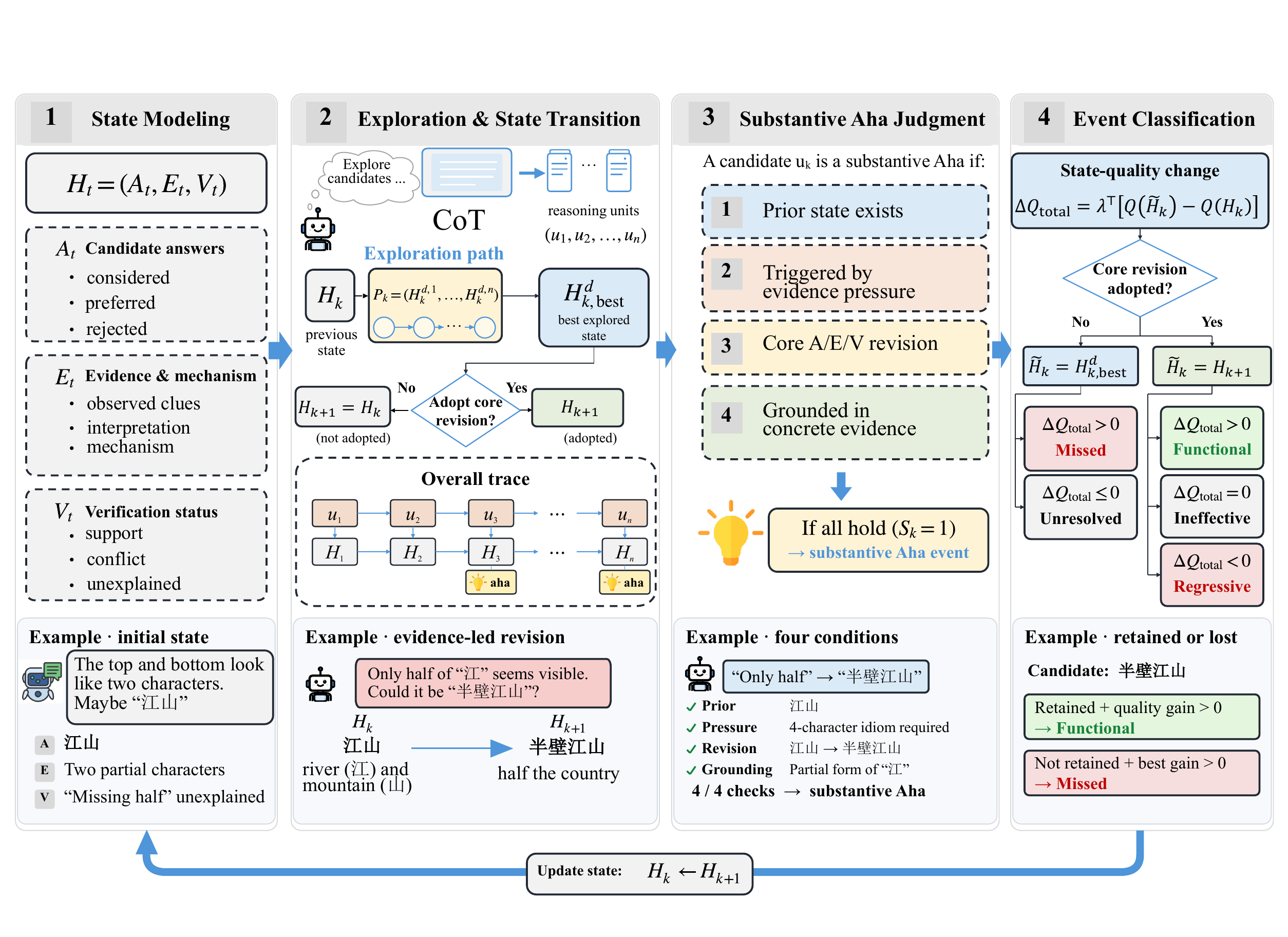}
    
    \caption{Overview of the AhaPro annotation framework:
    A/E/V state modeling, exploration and state transitions,
    substantive Aha judgment, and event classification.
    The bottom row presents a schematic example with alternative
    continuations illustrating Functional and Missed Aha.}
    \label{fig:rabench-overview}
\end{figure*}

\paragraph{Self-correction and interventions.}
Intrinsic self-correction can fail without external feedback \cite{huang2024selfcorrect}, while Tree of Thoughts explores and evaluates alternative reasoning paths \cite{yao2023tot}.
Failures after conflict detection can involve insufficient candidate discovery or the abandonment of an already discovered better hypothesis.
Interventions on intermediate CoT test whether reasoning steps influence subsequent predictions \cite{lanham2023faithfulness,zhao2025fakeaha}.
AhaPro constructs paired interventions from a shared conflict checkpoint, varying prefix retention and independent candidate search.
This design examines path dependence, restricted search, and failures to adopt better hypotheses.

\section{Benchmark Design}
\label{sec:benchmark-design}

\subsection{Dataset}
\label{sec:benchmark-construction}

\paragraph{Data collection.}
AhaPro collects puzzles from Baidu Wenku, Zhihu, Xiaohongshu,
and related studies, including \textit{Reasoning Riddles} and
\textit{Puzzled by Puzzles}. We manually remove items with
ambiguous answers, insufficient clues, or implausible
image--answer mappings. We then conduct further manual
screening to reduce the number of overly easy items.
The evaluation set contains 300 Chinese and 100 English puzzles.\footnote{The dataset is available at \url{https://anonymous.4open.science/r/AhaPro-25B7/}.}

\paragraph{Gold annotation.}
Each puzzle is annotated with a gold answer, the key visual
or textual evidence required to derive it, and the corresponding
interpretation mechanism. Two annotators independently label
each item, and disagreements are adjudicated by a third
annotator. The raw item-level agreement is 98.3\%.

\subsection{Framework Design}
\label{sec:process-annotation}

AhaPro tracks changes in reasoning states before and after
reflective behavior. Each CoT trace is segmented according to
semantic reasoning operations. A new unit boundary is introduced
when the model forms or changes a hypothesis, identifies or
reinterprets a clue, verifies a candidate answer, detects a
conflict, switches an interpretation mechanism, or adopts or
rejects a candidate answer. Consecutive statements serving the
same reasoning operation remain within the same unit.
Figure~\ref{fig:rabench-overview} summarizes the annotation
framework.

After each unit, the observable reasoning state is represented as
\begin{equation}
H_t = (\mathcal A_t,E_t,V_t).
\label{eq:reasoning-state}
\end{equation}
Here, $\mathcal A_t$ contains the current candidate answers and
their selection status; $E_t$ represents the evidence used;
and $V_t$ captures
relations of support, conflict, or unresolved correspondence
between candidate answers and evidence.

\paragraph{Aha moment identification.}
We use a two-stage protocol to identify Aha moments.
The first stage detects Aha candidates from observable signals,
including reflective expressions, failed verification, conflict
detection, candidate changes, mechanism changes, and the
introduction of previously unused evidence. The second stage
applies four criteria to identify substantive Aha events: a prior
reasoning state, an identifiable trigger, a core state revision,
and explicit evidence grounding:
\begin{equation}
S_k = C_k\,p_k\,t_k\,u_k\,g_k.
\label{eq:substantive-aha}
\end{equation}

For each substantive Aha, we further record whether exploration
reaches a better reasoning state and whether that state is
subsequently adopted. These two judgments, together with the
state-quality change, distinguish five outcomes: Missed,
Unresolved, Functional, Ineffective, and Regressive.

\paragraph{Annotation procedure.}
On 50 CoT traces, Qwen 3.8 Max and GPT-5.5 achieved
89.7\%, 87.2\%, and 94\% trace-level agreement on the
presence of substantive Aha, Functional Aha, and conflicts,
respectively, with an event-level F1 of 76.7\%.

\subsection{Diagnostic Metrics}
\label{sec:diagnostic-metrics}

\paragraph{State quality.}
We evaluate a reasoning state along three dimensions:
answer quality $q_A$, evidence coverage $q_E$, and
candidate--evidence consistency $q_V$:
\begin{equation}
Q(H) = [q_A(H), \, q_E(H), \, q_V(H)]^\top.
\label{eq:quality-vector}
\end{equation}
The three components respectively indicate whether the model
has reached the correct answer, identified the evidence required
by the puzzle, and used that evidence consistently with its
current hypothesis.

With equal weights
$\boldsymbol{\lambda}=(1/3,1/3,1/3)^{\top}$,
the quality change used for event classification is
\begin{equation}
\Delta Q_{\mathrm{total}}
=
\boldsymbol{\lambda}^{\top}
\left[Q(\widetilde H_k)-Q(H_k)\right].
\label{eq:quality-change}
\end{equation}
Here, $H_k$ is the pre-exploration state.
The comparison state $\widetilde H_k$ is the retained state
$H_{k+1}$ for adopted revisions and the best explored state
$H^d_{k,\mathrm{best}}$ otherwise.
Positive, zero, and negative values indicate improvement,
no net gain, and regression, respectively.

\paragraph{Exploration to gain.}
The first diagnostic chain tracks the progression from
Aha candidates to functional improvement. We report candidate-event density (CED),
substantive yield (SY), better-state discovery rate (BDR),
discovery adoption rate (DAR), and functional event density
(FED). These metrics form the following decomposition:
\begin{equation}
\mathrm{FED}_{10k} = \mathrm{CED}_{10k} \times \mathrm{SY} \times \mathrm{BDR} \times \mathrm{DAR}.
\label{eq:fed-decomposition}
\end{equation}

FED measures the frequency of functional revisions but does
not capture the magnitude or direction of state changes.
Let $\mathcal E$ index the unique substantive Aha events and $T$
denote the total number of CoT characters in the evaluated corpus.
We define normalized quality gain density as
\begin{equation}
\mathrm{NQGD}_{10k}
=\frac{10^4}{T}\sum_{k\in\mathcal E}
\boldsymbol{\lambda}^{\top}\left[Q(H_{k+1})-Q(H_k)\right].
\label{eq:nqgd}
\end{equation}
Here, $H_k$ and $H_{k+1}$ are the actual pre-event and post-event
states, respectively. NQGD summarizes their observed net quality
changes per 10,000 CoT characters.

\paragraph{Conflict-to-revision.}
The second diagnostic chain examines model behavior after
explicit conflict detection. CDR measures how often conflicts
are identified, CRR measures how often detected conflicts
lead to a Functional Aha, and FR captures cases in
which the model continues along its previous path without
achieving a functional revision.

Appendix~\ref{app:benchmark-details} provides the annotation protocol,
scoring rubrics, and metric definitions; annotation-agreement details
appear in Appendix B.2.
\section{Experiments and Results}
\label{sec:experiments-and-results} \definecolor{ahatableblue244}{RGB}{247,251,255} \definecolor{ahatableblue232}{RGB}{241,247,254} \definecolor{ahatableblue219}{RGB}{234,244,252} \definecolor{ahatableblue207}{RGB}{228,240,251} \definecolor{ahatableblue195}{RGB}{222,236,249} \definecolor{ahatableblue182}{RGB}{215,233,248} \definecolor{ahatableblue170}{RGB}{209,229,246} \definecolor{ahatableblue158}{RGB}{203,225,245} \definecolor{ahatableblue145}{RGB}{196,222,243} \definecolor{ahatableblue133}{RGB}{190,218,242}

\begin{table*}[!t]
\centering
\footnotesize
\setlength{\tabcolsep}{1.8pt}
\renewcommand{\arraystretch}{1.08}

\begin{tabularx}{\textwidth}
{@{}>{\raggedright\arraybackslash}X*{11}{c}@{}}

\toprule

& \multicolumn{2}{c}{\textbf{Overall}}
& \multicolumn{4}{c}{\textbf{Revision Funnel}}
& \multicolumn{2}{c}{\textbf{Functional Outcome}}
& \multicolumn{3}{c}{\textbf{Conflict Handling}} \\

\cmidrule(lr){2-3}
\cmidrule(lr){4-7}
\cmidrule(lr){8-9}
\cmidrule(lr){10-12}

\textbf{Model}
& \textbf{Acc.} $\uparrow$
& \textbf{Trace Len.}
& \textbf{CED}$_{10k}$
& \textbf{SY} $\uparrow$
& \textbf{BDR} $\uparrow$
& \textbf{DAR} $\uparrow$
& \textbf{FED}$_{10k}$ $\uparrow$
& \textbf{NQGD}$_{10k}$ $\uparrow$
& \textbf{CDR}
& \textbf{CRR} $\uparrow$
& \textbf{FR} $\downarrow$ \\

\midrule
\modelicon{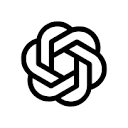}GPT-5.5 & \cellcolor{ahatableblue232}21.5 & \cellcolor{ahatableblue244}153.9 & \cellcolor{ahatableblue219}22.909 & \cellcolor{ahatableblue158}56.0 & \cellcolor{ahatableblue244}49.4 & \cellcolor{ahatableblue133}92.3 & \cellcolor{ahatableblue207}5.849 & \cellcolor{ahatableblue219}0.988 & \cellcolor{ahatableblue244}15.8 & \cellcolor{ahatableblue207}46.0 & \cellcolor{ahatableblue232}14.3 \\
\modelicon{openai}GPT-5.6 Sol & \cellcolor{ahatableblue232}19.5 & \cellcolor{ahatableblue244}104.8 & \cellcolor{ahatableblue207}27.675 & \cellcolor{ahatableblue133}67.2 & \cellcolor{ahatableblue244}47.4 & \cellcolor{ahatableblue133}89.2 & \cellcolor{ahatableblue195}7.873 & \cellcolor{ahatableblue195}1.690 & \cellcolor{ahatableblue244}11.2 & \cellcolor{ahatableblue195}48.9 & \cellcolor{ahatableblue244}11.1 \\
\modelicon{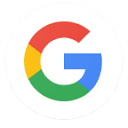}Gemini 2.5 Flash & \cellcolor{ahatableblue232}19.5 & \cellcolor{ahatableblue232}2736.9 & \cellcolor{ahatableblue232}17.337 & \cellcolor{ahatableblue219}26.0 & \cellcolor{ahatableblue133}73.3 & \cellcolor{ahatableblue158}79.0 & \cellcolor{ahatableblue232}2.612 & \cellcolor{ahatableblue232}0.653 & \cellcolor{ahatableblue170}52.0 & \cellcolor{ahatableblue170}54.3 & \cellcolor{ahatableblue182}23.1 \\
\modelicon{google}Gemini 3.5 Flash & \cellcolor{ahatableblue158}47.8 & \cellcolor{ahatableblue244}1492.1 & \cellcolor{ahatableblue232}14.124 & \cellcolor{ahatableblue207}34.5 & \cellcolor{ahatableblue195}59.1 & \cellcolor{ahatableblue145}86.0 & \cellcolor{ahatableblue232}2.480 & \cellcolor{ahatableblue232}0.810 & \cellcolor{ahatableblue207}38.0 & \cellcolor{ahatableblue182}50.7 & \cellcolor{ahatableblue244}13.2 \\
\modelicon{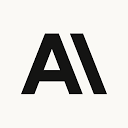}Claude Opus 4.8 & \cellcolor{ahatableblue207}27.5 & \cellcolor{ahatableblue244}310.0 & \cellcolor{ahatableblue133}69.283 & \cellcolor{ahatableblue158}52.0 & \cellcolor{ahatableblue219}53.2 & \cellcolor{ahatableblue145}82.8 & \cellcolor{ahatableblue133}15.889 & \cellcolor{ahatableblue133}3.777 & \cellcolor{ahatableblue182}46.5 & \cellcolor{ahatableblue207}46.2 & \cellcolor{ahatableblue232}14.5 \\
\modelicon{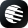}Kimi K2.6 & \cellcolor{ahatableblue207}31.2 & \cellcolor{ahatableblue232}2728.6 & \cellcolor{ahatableblue219}24.041 & \cellcolor{ahatableblue244}14.7 & \cellcolor{ahatableblue158}68.1 & \cellcolor{ahatableblue207}57.3 & \cellcolor{ahatableblue244}1.374 & \cellcolor{ahatableblue244}0.451 & \cellcolor{ahatableblue170}56.8 & \cellcolor{ahatableblue244}35.7 & \cellcolor{ahatableblue133}31.3 \\
\modelicon{moonshot}Kimi K3 & \cellcolor{ahatableblue158}48.2 & \cellcolor{ahatableblue133}19204.5 & \cellcolor{ahatableblue244}5.553 & \cellcolor{ahatableblue219}27.1 & \cellcolor{ahatableblue232}51.3 & \cellcolor{ahatableblue182}67.1 & \cellcolor{ahatableblue244}0.518 & \cellcolor{ahatableblue244}0.183 & \cellcolor{ahatableblue133}78.3 & \cellcolor{ahatableblue195}47.3 & \cellcolor{ahatableblue133}32.5 \\
\modelicon{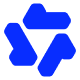}Qwen 3.7 Plus & \cellcolor{ahatableblue182}36.2 & \cellcolor{ahatableblue232}3846.6 & \cellcolor{ahatableblue232}15.774 & \cellcolor{ahatableblue207}34.8 & \cellcolor{ahatableblue207}57.5 & \cellcolor{ahatableblue195}61.3 & \cellcolor{ahatableblue244}1.937 & \cellcolor{ahatableblue232}0.712 & \cellcolor{ahatableblue133}77.8 & \cellcolor{ahatableblue195}47.6 & \cellcolor{ahatableblue195}20.9 \\
\modelicon{qwen}Qwen 3.8 Max & \cellcolor{ahatableblue133}57.2 & \cellcolor{ahatableblue244}1311.0 & \cellcolor{ahatableblue232}16.476 & \cellcolor{ahatableblue182}44.3 & \cellcolor{ahatableblue195}59.8 & \cellcolor{ahatableblue170}75.5 & \cellcolor{ahatableblue232}3.299 & \cellcolor{ahatableblue219}1.157 & \cellcolor{ahatableblue207}38.0 & \cellcolor{ahatableblue133}67.3 & \cellcolor{ahatableblue170}26.0 \\
\modelicon{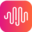}MiniMax M3 & \cellcolor{ahatableblue244}14.2 & \cellcolor{ahatableblue244}1966.4 & \cellcolor{ahatableblue219}19.452 & \cellcolor{ahatableblue182}43.0 & \cellcolor{ahatableblue219}52.6 & \cellcolor{ahatableblue170}76.0 & \cellcolor{ahatableblue232}3.344 & \cellcolor{ahatableblue232}0.818 & \cellcolor{ahatableblue195}42.2 & \cellcolor{ahatableblue244}33.1 & \cellcolor{ahatableblue170}25.4 \\
\modelicon{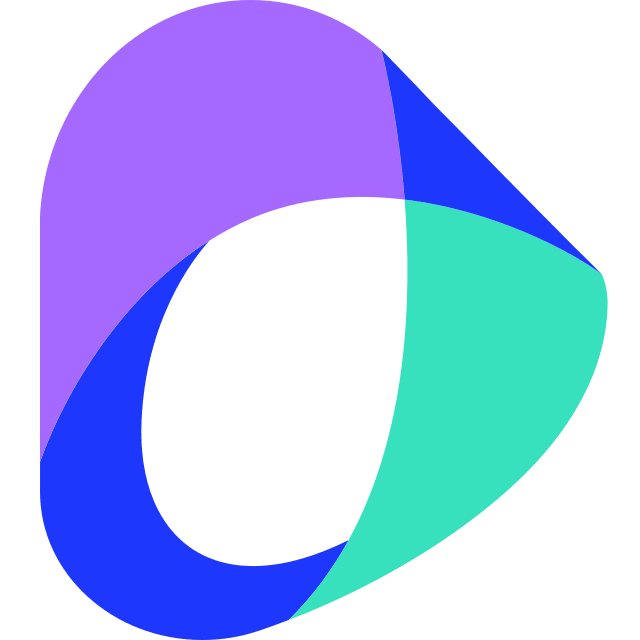}Doubao 2.1 Pro & \cellcolor{ahatableblue170}41.2 & \cellcolor{ahatableblue158}13898.1 & \cellcolor{ahatableblue244}6.470 & \cellcolor{ahatableblue182}43.6 & \cellcolor{ahatableblue145}69.0 & \cellcolor{ahatableblue244}39.8 & \cellcolor{ahatableblue244}0.775 & \cellcolor{ahatableblue244}0.303 & \cellcolor{ahatableblue158}62.8 & \cellcolor{ahatableblue158}59.0 & \cellcolor{ahatableblue232}13.5 \\
\bottomrule
\end{tabularx}
\caption{
Main results of task performance and diagnostics of Aha moments
across models.
Darker blue indicates larger values in 10 equal-width intervals per numeric column in Tables~\ref{tab:main_results}--\ref{tab:discovery-retention-gap}; confidence intervals are unshaded.
Acc., SY, BDR, DAR, CDR, CRR and FR are percentages.
The evaluation set contains 400 puzzles;
missing responses and unavailable final answers count as incorrect.
Trace Len. is recorded CoT characters per assigned puzzle.
Process metrics use observable CoT, including partial traces.
}
\label{tab:main_results}

\end{table*}

This section evaluates Aha moments in the reasoning of different models
under AhaPro. We first compare their final task performance and then
analyze the relationships among candidate exploration, state improvement,
candidate adoption, and effective revision using the
exploration-to-gain and conflict-to-revision metric suites. Definitions
and computation of the process-level metrics are provided in
Section~\ref{sec:diagnostic-metrics}.

\subsection{Experimental Setup}
\label{sec:experimental-setup}

We evaluate proprietary and open multimodal models from different model families. All models receive the original puzzle image and semantically equivalent task instructions through their native chat templates and visible-reasoning interfaces. Each model solves the puzzle independently and returns a final answer. Final-answer accuracy is measured against gold answers. Process evaluation reconstructs reasoning-state sequences following Section~\ref{sec:process-annotation} and reports the metrics defined in Section~\ref{sec:diagnostic-metrics}. Experimental configurations and supplementary analyses are documented in Appendices~\ref{app:experimental-details} and~\ref{app:answer-transitions}--\ref{app:process-counts}.

\subsection{Main Results}
\label{sec:answer-aha}

Table~\ref{tab:main_results} summarizes the results of 11 models on 400 puzzles, with final-answer accuracy ranging from 14.2\% to 57.2\%.

To examine how revisions during reasoning relate to answer-level gains, we compare the initial and final answers within each CoT.
Table~\ref{tab:answer-aha} summarizes these answer transitions and their relationship with Functional Aha.
On paired observations, all models show higher final than initial answer accuracy.
Comparing $R$ and Aha-LCR further shows that a substantial share of wrong-to-correct answer transitions is accompanied by Functional Aha.

\begin{table}[t]
\centering
\footnotesize
\setlength{\tabcolsep}{2pt}
\renewcommand{\arraystretch}{1.08}

\begin{tabular*}{\columnwidth}{@{\extracolsep{\fill}}lcccccc@{}}
\toprule
Model & $I$ & $R$ & $D$ & $F$ & $\Delta$ & Aha-LCR \\
\midrule
GPT-5.5 & \cellcolor{ahatableblue244}8.1 & \cellcolor{ahatableblue232}5.9 & \cellcolor{ahatableblue244}0.0 & \cellcolor{ahatableblue244}13.5 & \cellcolor{ahatableblue195}5.4 & \cellcolor{ahatableblue244}2.9 \\
GPT-5.6 Sol & \cellcolor{ahatableblue244}4.5 & \cellcolor{ahatableblue195}11.9 & \cellcolor{ahatableblue244}0.0 & \cellcolor{ahatableblue244}15.9 & \cellcolor{ahatableblue133}11.4 & \cellcolor{ahatableblue219}7.1 \\
Gemini 2.5 Flash & \cellcolor{ahatableblue232}16.8 & \cellcolor{ahatableblue244}3.1 & \cellcolor{ahatableblue244}0.0 & \cellcolor{ahatableblue244}19.4 & \cellcolor{ahatableblue232}2.6 & \cellcolor{ahatableblue244}3.1 \\
Gemini 3.5 Flash & \cellcolor{ahatableblue195}42.2 & \cellcolor{ahatableblue244}4.0 & \cellcolor{ahatableblue207}4.1 & \cellcolor{ahatableblue195}42.8 & \cellcolor{ahatableblue244}0.6 & \cellcolor{ahatableblue244}3.0 \\
Claude Opus 4.8 & \cellcolor{ahatableblue232}14.6 & \cellcolor{ahatableblue232}6.5 & \cellcolor{ahatableblue195}5.4 & \cellcolor{ahatableblue244}19.4 & \cellcolor{ahatableblue207}4.7 & \cellcolor{ahatableblue232}5.6 \\
Kimi K2.6 & \cellcolor{ahatableblue207}29.2 & \cellcolor{ahatableblue244}4.1 & \cellcolor{ahatableblue182}5.7 & \cellcolor{ahatableblue219}30.4 & \cellcolor{ahatableblue244}1.2 & \cellcolor{ahatableblue244}3.5 \\
Kimi K3 & \cellcolor{ahatableblue182}52.5 & \cellcolor{ahatableblue195}11.8 & \cellcolor{ahatableblue195}5.3 & \cellcolor{ahatableblue182}55.3 & \cellcolor{ahatableblue219}2.8 & \cellcolor{ahatableblue195}11.8 \\
Qwen 3.7 Plus & \cellcolor{ahatableblue207}31.6 & \cellcolor{ahatableblue207}9.9 & \cellcolor{ahatableblue207}4.3 & \cellcolor{ahatableblue207}37.0 & \cellcolor{ahatableblue195}5.4 & \cellcolor{ahatableblue207}9.4 \\
Qwen 3.8 Max & \cellcolor{ahatableblue182}50.0 & \cellcolor{ahatableblue133}22.8 & \cellcolor{ahatableblue182}5.9 & \cellcolor{ahatableblue170}58.4 & \cellcolor{ahatableblue158}8.4 & \cellcolor{ahatableblue133}21.8 \\
MiniMax M3 & \cellcolor{ahatableblue244}10.5 & \cellcolor{ahatableblue244}3.9 & \cellcolor{ahatableblue133}11.1 & \cellcolor{ahatableblue244}12.8 & \cellcolor{ahatableblue232}2.3 & \cellcolor{ahatableblue244}3.5 \\
Doubao 2.1 Pro & \cellcolor{ahatableblue133}85.6 & \cellcolor{ahatableblue170}15.8 & \cellcolor{ahatableblue232}1.8 & \cellcolor{ahatableblue133}86.4 & \cellcolor{ahatableblue244}0.8 & \cellcolor{ahatableblue195}10.5 \\
\bottomrule
\end{tabular*}
\caption{Answer-layer diagnostics of Aha moments. $I$: initial accuracy; $R$: correction given an initial error; $D$: degradation given an initially correct answer; $F$: paired final accuracy. Rates are percentages; $\Delta=F-I$ is in percentage points.}
\label{tab:answer-aha}
\par\smallskip
\begin{minipage}{\columnwidth}\footnotesize\raggedright
Aha-LCR: Aha-linked correction rate (Appendix~\ref{app:answer-transitions}). $F$ uses the paired subset.
-- denotes undefined or unavailable values.
\end{minipage}
\end{table}

Process-level metrics reveal distinct bottlenecks in better-state discovery and adoption. Final-answer accuracy alone does not identify these bottlenecks:

\textbf{First, similar final performance can correspond to different revision processes.}
GPT-5.6 Sol and Gemini 2.5 Flash both achieve 19.5\% accuracy, but differ in substantive yield (67.2\% versus 26.0\%) and better-state discovery (47.4\% versus 73.3\%).
GPT-5.6 Sol's candidate events more often qualify as substantive revisions, while Gemini 2.5 Flash's substantive revisions more often discover better states.

\begingroup
\clubpenalty=10000
\textbf{Second, substantial Aha gains are lost between better-state discovery and adoption.}
Doubao 2.1 Pro and Kimi K2.6 have similar BDRs (69.0\% and 68.1\%), yet their DARs are 39.8\% and 57.3\%, respectively.
Doubao misses 653 of the 1,084 better states it discovers.
These events identify better explored states whose core content is not retained in subsequent reasoning.
\par
\endgroup

\textbf{Third, a substantial revision gap remains after conflict detection.}
Qwen 3.7 Plus detects conflicts in 77.8\% of observable traces and revises effectively in only 47.6\% of those cases.
Qwen 3.8 Max detects conflicts in 38.0\% of observable traces, with a CRR of 67.3\%.
For Qwen 3.7 Plus, over half of the traces with explicitly detected conflicts do not contain a functional revision after conflict detection.
This places failure after the problem has already been recognized.
Its high conflict-detection rate and lower conflict-to-revision rate jointly characterize this bottleneck.

\begin{figure}[!t]
    \centering
    \includegraphics[width=\columnwidth]
        {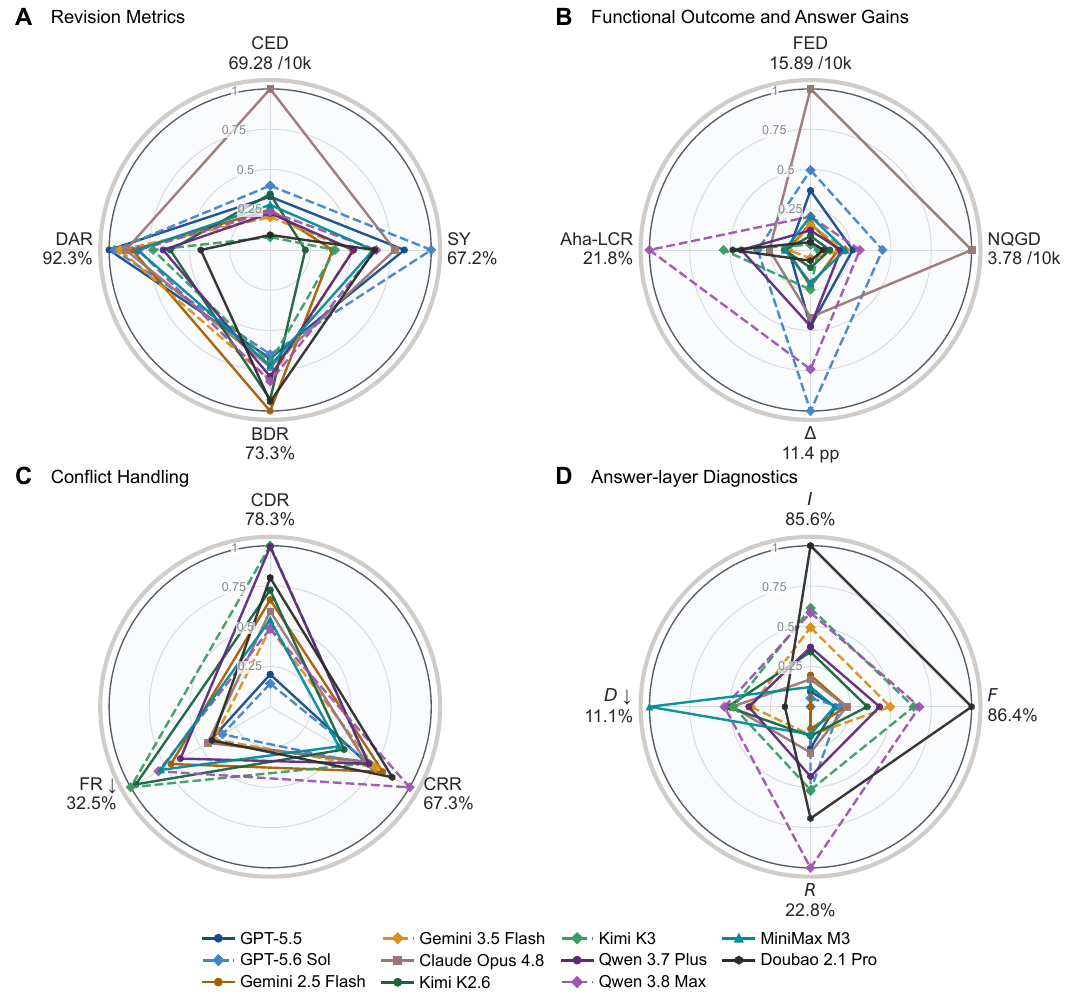}
    \caption{
        Diagnostics of Aha moments across models:
        (A) revision metrics;
        (B) functional outcomes and answer gains;
        (C) conflict handling;
        (D) answer-layer diagnostics.
        Each axis is scaled by its across-model maximum,
        shown in the original units.
        Density metrics are per 10,000 CoT characters.
        Lower FR and $D$ are preferable.
        $\Delta$, Aha-LCR, and all metrics in (D) use
        model-specific paired subsets; $F$ differs from
        full-set accuracy in Table~\ref{tab:main_results}.
    }
    \label{fig:aha-diagnostics-overview}
\end{figure}

Figure~\ref{fig:aha-diagnostics-overview} provides an overview of the above process-level and answer-layer diagnostics.

\subsection{Process-level Analysis}
\label{sec:process-analysis}

\begin{figure}[!t]
    \centering
    \includegraphics[width=\columnwidth]{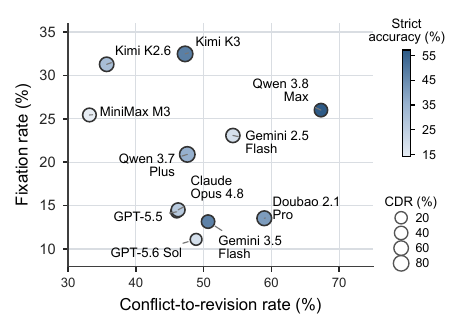}
    \caption{Conflict handling across models, showing how detected conflicts are converted into effective revisions.}
    \label{fig:conflict-handling}
    \label{fig:process-analysis}
\end{figure}

\begin{figure}[!t]
    \centering
    \includegraphics[width=\columnwidth]{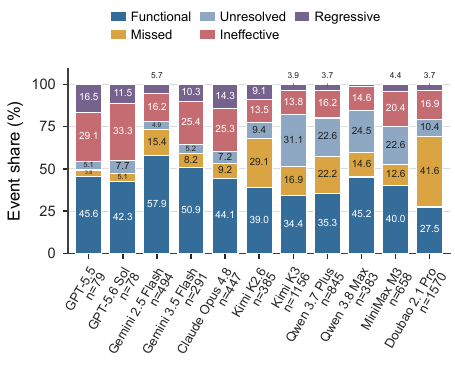}
    \caption{Distribution of substantive Aha outcomes across eleven models.}
    \label{fig:aha-outcomes}
\end{figure}

To further examine where revision succeeds or fails, we analyze the
transition from better-state discovery to adoption, the conversion from
conflict detection to effective revision, the distribution of Aha
outcomes, and the component-wise changes in reasoning-state quality.
Figures~\ref{fig:conflict-handling} and~\ref{fig:aha-outcomes} summarize conflict handling and Aha outcomes.

\paragraph{Better-state discovery does not guarantee adoption.}
Table~\ref{tab:discovery-retention-gap} measures the gap between the best
previously attained and retained state quality at the last scorable
substantive-Aha checkpoint. Positive gaps occur in all model-specific
subsets: 37.3\% of Doubao 2.1 Pro cases and 24.8\% of Qwen 3.7 Plus
cases, with mean gaps of 0.171 and 0.075. These gaps capture incomplete
adoption or later regression within the trace.

\begin{table}[!t]
\centering
\footnotesize
\setlength{\tabcolsep}{2pt}
\renewcommand{\arraystretch}{1.12}

\begin{tabularx}{\columnwidth}
{@{}>{\raggedright\arraybackslash}Xccc>{\scriptsize}cc@{}}

\toprule

\textbf{Model}
& \textbf{Best}
& \textbf{Ret.}
& \textbf{Gap}
& \textbf{95\% CI}
& \textbf{$g_i>0$} \\

\midrule
GPT-5.5 & \cellcolor{ahatableblue244}0.443 & \cellcolor{ahatableblue244}0.394 & \cellcolor{ahatableblue232}0.049 & [0.023, 0.079] & \cellcolor{ahatableblue195}21.4 \\
GPT-5.6 Sol & \cellcolor{ahatableblue244}0.476 & \cellcolor{ahatableblue232}0.440 & \cellcolor{ahatableblue244}0.036 & [0.014, 0.066] & \cellcolor{ahatableblue207}20.4 \\
Gemini 2.5 Flash & \cellcolor{ahatableblue232}0.498 & \cellcolor{ahatableblue219}0.475 & \cellcolor{ahatableblue244}0.023 & [0.014, 0.033] & \cellcolor{ahatableblue244}10.3 \\
Gemini 3.5 Flash & \cellcolor{ahatableblue195}0.623 & \cellcolor{ahatableblue182}0.583 & \cellcolor{ahatableblue232}0.040 & [0.023, 0.059] & \cellcolor{ahatableblue219}16.9 \\
Claude Opus 4.8 & \cellcolor{ahatableblue219}0.544 & \cellcolor{ahatableblue219}0.483 & \cellcolor{ahatableblue219}0.061 & [0.046, 0.077] & \cellcolor{ahatableblue145}32.4 \\
Kimi K2.6 & \cellcolor{ahatableblue195}0.637 & \cellcolor{ahatableblue207}0.518 & \cellcolor{ahatableblue170}0.118 & [0.088, 0.151] & \cellcolor{ahatableblue133}34.9 \\
Kimi K3 & \cellcolor{ahatableblue145}0.782 & \cellcolor{ahatableblue158}0.668 & \cellcolor{ahatableblue170}0.114 & [0.094, 0.136] & \cellcolor{ahatableblue145}32.4 \\
Qwen 3.7 Plus & \cellcolor{ahatableblue170}0.701 & \cellcolor{ahatableblue170}0.625 & \cellcolor{ahatableblue207}0.075 & [0.058, 0.094] & \cellcolor{ahatableblue182}24.8 \\
Qwen 3.8 Max & \cellcolor{ahatableblue133}0.804 & \cellcolor{ahatableblue133}0.747 & \cellcolor{ahatableblue219}0.057 & [0.036, 0.081] & \cellcolor{ahatableblue219}16.7 \\
MiniMax M3 & \cellcolor{ahatableblue219}0.533 & \cellcolor{ahatableblue219}0.483 & \cellcolor{ahatableblue232}0.049 & [0.034, 0.066] & \cellcolor{ahatableblue219}17.5 \\
Doubao 2.1 Pro & \cellcolor{ahatableblue133}0.833 & \cellcolor{ahatableblue158}0.662 & \cellcolor{ahatableblue133}0.171 & [0.145, 0.198] & \cellcolor{ahatableblue133}37.3 \\
\bottomrule
\end{tabularx}
\caption{
Discovery--retention quality gaps at the last scorable
substantive-Aha checkpoint.
Best and Ret. are mean state qualities;
95\% CI gives the confidence interval for the mean gap.
All statistics use the model-specific scorable subsets.
Zero gaps remain in the means and denominators.
}
\label{tab:discovery-retention-gap}

\end{table}

\paragraph{Models differ in revision and fixation after conflict detection.}
Figure~\ref{fig:conflict-handling} shows different post-conflict behaviors
despite similar conflict-detection rates for Doubao 2.1 Pro and Kimi K2.6.
Doubao more often converts detected conflicts into functional revisions
(CRR: 59.0\% versus 35.7\%) and less often retains the previous path
(FR: 13.5\% versus 31.3\%).
These differences show that post-conflict behavior further distinguishes
models' revision capabilities.

\paragraph{Models exhibit different revision failure modes.}
Figure~\ref{fig:aha-outcomes} shows that substantive Aha events can fail to yield functional improvement in different ways.
Doubao 2.1 Pro records 653 Missed events among 1,570 substantive events (41.6\%).
Its large Missed share and low DAR indicate a bottleneck in adopting better explored states.
GPT-5.5 records 23 Ineffective and 13 Regressive events among 79 substantive events.
In these cases, revisions are adopted but leave state quality unchanged or reduce it.
These outcomes distinguish failures to retain the best explored state's core content from ineffective or regressive adopted revisions.

\paragraph{Aha moments involve changes beyond the final answer.}
The A/E/V decomposition shows that models differ in their relative gains
in answer quality, evidence coverage, and candidate--evidence consistency.
Evidence use and consistency can improve without an immediate correct
answer.

\paragraph{Human-evaluated ablation.}
On 110 CoT traces with 62 human-confirmed effective revision events,
the full A/E/V criteria achieve 80.3\% event-level F1, compared with
65.3\% for simple reflection cues. Removing answer quality, evidence
coverage, or candidate--evidence consistency reduces F1 to 74.4\%,
62.6\%, and 64.2\%, respectively.

\subsection{Aha Moment Dynamics across Languages and Puzzle Categories}
\label{sec:language_category_analysis}

\begin{table}[!t]
\centering
\footnotesize
\setlength{\tabcolsep}{2pt}
\renewcommand{\arraystretch}{1.12}

\begin{tabularx}{\columnwidth}
{@{}>{\raggedright\arraybackslash}Xrr@{}}
\toprule
\multicolumn{3}{@{}l@{}}{
    \textbf{(a) Overall language comparison}
} \\
\midrule

Metric
& \shortstack[r]{Chinese\\($n=300$)}
& \shortstack[r]{English\\($n=100$)} \\

\midrule
Candidate occurrence (\%)
& 66.0 & 70.6 \\

Candidates per CoT
& 3.85 & 5.91 \\

Substantive Aha occurrence (\%)
& 55.8 & 47.4 \\

SY (\%)
& 47.5 & 21.1 \\

CDR (\%)
& 45.2 & 53.3 \\

CRR (\%)
& 53.7 & 39.4 \\

Paired answer correction (\%)
& 11.1 & 2.5 \\

\bottomrule
\end{tabularx}

\vspace{3pt}

\begin{tabularx}{\columnwidth}
{@{}l>{\raggedright\arraybackslash}Xrrr@{}}
\toprule
\multicolumn{5}{@{}l@{}}{
    \textbf{(b) Within-language category comparisons}
} \\
\midrule

Language
& Mechanism
& $n^{+}/n^{-}$
& \shortstack[r]{Candidate\\(\%)}
& \shortstack[r]{Substantive\\Aha (\%)} \\

\midrule
Chinese
& Absence/\newline negation
& 21/279
& 73.2 / 65.5
& 66.2 / 55.0 \\

Chinese
& Character\newline manipulation
& 27/273
& 69.0 / 65.7
& 63.6 / 55.0 \\

English
& Image\newline recognition
& 79/21
& 72.6 / 63.2
& 49.0 / 41.1 \\

\bottomrule
\end{tabularx}
\caption{Aha moment dynamics across languages and puzzle mechanisms.}
\label{tab:language_category}

\vspace{3pt}

\begin{minipage}{\columnwidth}
\scriptsize
\raggedright
Results are averaged equally across models.
Process rates use observable CoT.
Paired correction uses initially incorrect traces with
observable initial and final answers.
In (b), $n^{+}/n^{-}$ and rate pairs refer to puzzles
with/without the mechanism label; labels are non-exclusive.
Occurrence rates measure traces containing at least one
corresponding event.
Comparisons between languages are observational
associations within our corpora.
\end{minipage}

\end{table}

Overall, Table~\ref{tab:language_category} shows differences
in Aha candidate occurrence, substantive restructuring,
and subsequent revision across languages and puzzle mechanisms.

First, English traces contain more Aha candidates and show
more frequent conflict detection, alongside a lower substantive
yield. SY is 21.1\% for English and 47.5\% for Chinese,
with lower English SY across all models. English also exhibits
lower rates of effective revision after conflict and answer
correction among paired traces.

Second, puzzle mechanisms are associated with differences
in candidate occurrence and substantive restructuring.
In Chinese puzzles, absence/negation and character manipulation
have substantive Aha occurrence rates 11.2 and 8.6 percentage
points higher than their respective comparison groups.
In English puzzles, image recognition is associated with
a 9.4 percentage point higher Aha candidate occurrence rate.
Missing information and character manipulation may prompt
reinterpretation of clues. Mapping images to words may
introduce more competing candidates.

\section{Intervention}
\label{sec:intervention}

\subsection{History and Candidate Selection}

Our preceding analysis shows that some models fail to revise effectively after detecting conflicts and also abandon better states they have discovered.
This raises two questions: Does prior reasoning context hinder revision after conflict detection?
How do candidate search and subsequent selection each limit effective revision?
A paired $2\times2$ design compares retaining or removing prior reasoning
(P1/P0), and single-path continuation or three independent rollouts
followed by selection (S1/SB). P0 retains the image, task, and standardized
evidence/conflict information. Screening all 400 puzzles yields complete
four-condition data for 60 incorrect checkpoints in Qwen 3.7 Plus (Q3.7)
and 44 in Qwen 3.8 Max (Q3.8), plus 22 and 31 correct controls, respectively.

\begin{table}[t]
\centering
\small
\setlength{\tabcolsep}{3pt}

\begin{tabular*}{\columnwidth}{@{\extracolsep{\fill}}llrrrr@{}}
\toprule
& & \multicolumn{2}{c}{S1 (\%)} & \multicolumn{2}{c}{SB} \\
\cmidrule(lr){3-4}\cmidrule(l){5-6}
Model & Hist. & Ret.$\downarrow$ & AIR$\uparrow$ & Found & Adopted \\
\midrule
Q3.7 & P1 & 56.7 & 10.0 & 8/60 & 7/8 \\
 & P0 & 25.0 & 18.3 & 16/60 & 10/16 \\
\midrule
Q3.8 & P1 & 47.7 & 11.4 & 9/44 & 9/9 \\
 & P0 & 9.1 & 15.9 & 9/44 & 9/9 \\
\bottomrule
\end{tabular*}
\caption{Paired intervention results on incorrect checkpoint candidates.
Ret.\ is original-candidate retention; Ret.\ and AIR are percentages.
Found reports gold-containing pools/items; Adopted reports
correct selections/gold-containing pools ($K=3$).}
\label{tab:intervention_results}
\end{table}

Removing history reduced wrong-candidate retention by 31.7 and 38.6
percentage points for Q3.7 and Q3.8, respectively
(Table~\ref{tab:intervention_results}; both Holm-adjusted $p<.001$).
Answer repair rates increased slightly. Under P0-SB, Q3.7 doubled correct-answer discovery relative to P1-SB,
but selected only 10 of 16 correct answers.
Reset also increased observed errors on correct
controls, suggesting that context removal can disrupt successful
interpretations.

These exploratory replays estimate the combined effect of context
changes; SB also uses more tokens.

\subsection{Incremental Information Assistance}
\label{sec:information-assistance}

The discovery of better states during Aha moments may be limited by visual clue recognition and the selection of interpretation mechanisms.
To assess how information support affects answer accuracy and the subsequent discovery and adoption of better states,
we compare image-only input (T0), added observation facts (T1), and facts plus a mechanism cue (T2) on the same 100 puzzles per model.
Supplied information is excluded from spontaneous Aha discovery.

\begin{table}[t]
\centering
\footnotesize
\setlength{\tabcolsep}{3pt}
\renewcommand{\arraystretch}{1.10}

\begin{tabular*}{\columnwidth}{@{\extracolsep{\fill}}lcrrr@{}}
\toprule
\textbf{Model}
& \textbf{Input}
& \textbf{Acc.}
& \textbf{BDR}
& \textbf{DAR} \\
\midrule
Qwen 3.7 Plus
& T0 & 47.0 & 50.3\textsuperscript{*}
& 57.7\textsuperscript{*} \\
& T1 & 61.0 & 81.4 & 74.7 \\
& T2 & 66.0 & 73.9 & 80.5 \\
\midrule
Qwen 3.8 Max
& T0 & 63.0 & 61.0\textsuperscript{*}
& 76.6\textsuperscript{*} \\
& T1 & 73.0 & 75.9 & 78.8 \\
& T2 & 74.0 & 81.3 & 78.5 \\
\midrule
Opus 4.8
& T0 & 39.0 & 55.6\textsuperscript{*}
& 87.3\textsuperscript{*} \\
& T1 & 55.0 & 90.0\textsuperscript{\ddag}
& 88.9\textsuperscript{\ddag} \\
& T2 & 53.0 & 69.2\textsuperscript{\S}
& 83.3\textsuperscript{\S} \\
\bottomrule
\end{tabular*}
\caption{Incremental information assistance on 100 puzzles per model
and condition. All metrics are percentages.}
\label{tab:information-assistance}

\par\smallskip
\begin{minipage}{\columnwidth}
\footnotesize\raggedright
T0: image; T1: image + facts;
T2: image + the same facts + mechanism cue.
Annotation sources: * historical T0;
\ddag\ Opus T1 pilot; \S\ Opus T2 review.
Opus T1/T2 include 13/30 and 7/26 boundary-uncertain events,
respectively.
\end{minipage}
\end{table}

In Table~\ref{tab:information-assistance}, observation facts improve
accuracy by 14, 10, and 16 percentage points for the two Qwen models
and Opus, respectively. Mechanism cues add $+5$, $+1$, and $-2$ points,
showing mixed further gains. Process comparisons retain annotation
differences.

\begin{samepage}
Intervention prompts, implementation details, and supplementary results
are provided in Appendices~\ref{app:prompts} and~\ref{app:history-intervention}--\ref{app:information-details}.\par
\end{samepage}

\section{Conclusion}
\label{sec:conclusion}

We introduced AhaPro, a rebus benchmark with a process-level framework
that separates Aha candidates from substantive revisions and tracks
answer, evidence, and verification states. Across the evaluated
CoT traces from eleven models, conflict recognition, better-state
discovery, and subsequent adoption showed distinct failure patterns.
Exploratory interventions reduced incorrect-candidate retention after
context reset. Multi-branch search further highlighted answer selection
as a bottleneck between correct-answer discovery and successful correction. These analyses locate where revision succeeds or fails
and connect intermediate state changes with downstream answer outcomes.

\section*{Limitations}
\label{sec:limitations}

AhaPro evaluates observable reasoning-state changes in generated CoT text. These observations cannot establish whether genuine Aha moments occur internally. Failure to detect Aha in the text does not imply its absence internally. Public sources and subjective judgments in manual screening and annotation review may affect puzzle distribution, event labeling, and state-quality scores. Our findings characterize the selected puzzles and evaluated CoT. Answer correctness alone cannot establish whether puzzles appeared in model training data. Aha triggers and behavioral characteristics may vary with post-training methods, which this study does not separately assess. Future work could investigate Aha mechanisms through model architectures and post-training designs. Chinese and English puzzles with matched solution mechanisms and language-specific items could clarify candidate revision and better-state discovery and adoption across languages.


\bibliography{Reference}

\begin{thebibliography}{30}
\providecommand{\natexlab}[1]{#1}

\bibitem[{Chen et~al.(2025)Chen, Xu, Liang, He, Pang, Yu, Song, Liu, Zhou,
  Zhang, Wang, Tu, Mi, and Yu}]{chen2025overthinking}
Xingyu Chen, Jiahao Xu, Tian Liang, Zhiwei He, Jianhui Pang, Dian Yu, Linfeng
  Song, Qiuzhi Liu, Mengfei Zhou, Zhuosheng Zhang, Rui Wang, Zhaopeng Tu,
  Haitao Mi, and Dong Yu. 2025.
\newblock \href {https://proceedings.mlr.press/v267/chen25bx.html} {Do {NOT}
  think that much for 2+3=? {On} the overthinking of long reasoning models}.
\newblock In \emph{Proceedings of the 42nd International Conference on Machine
  Learning}, volume 267 of \emph{Proceedings of Machine Learning Research},
  pages 9487--9499. PMLR.

\bibitem[{d'Aliberti and Ribeiro(2026)}]{daliberti2026illusion}
Liv~G. d'Aliberti and Manoel~Horta Ribeiro. 2026.
\newblock \href {https://doi.org/10.18653/v1/2026.findings-acl.47} {The
  illusion of insight in reasoning models}.
\newblock In \emph{Findings of the Association for Computational Linguistics:
  ACL 2026}, pages 924--966. Association for Computational Linguistics.

\bibitem[{Das et~al.(2025)Das, Nandy, Bajaj, and S}]{das2025rebus}
Trishanu Das, Abhilash Nandy, Khush Bajaj, and Deepiha S. 2025.
\newblock \href {https://arxiv.org/abs/2511.01340}
  {$\left|\,\circlearrowright\,\boxed{\text{bus}}\,\right|$: A large and
  diverse multimodal benchmark for evaluating the ability of vision-language
  models to understand rebus puzzles}.
\newblock \emph{Preprint}, arXiv:2511.01340.

\bibitem[{{DeepSeek-AI}(2025)}]{deepseek2025r1}
{DeepSeek-AI}. 2025.
\newblock \href {https://arxiv.org/abs/2501.12948} {{DeepSeek-R1}:
  Incentivizing reasoning capability in llms via reinforcement learning}.
\newblock \emph{Preprint}, arXiv:2501.12948.

\bibitem[{Golovneva et~al.(2023)Golovneva, Chen, Poff, Corredor, Zettlemoyer,
  Fazel-Zarandi, and Celikyilmaz}]{golovneva2023roscoe}
Olga Golovneva, Moya Chen, Spencer Poff, Martin Corredor, Luke Zettlemoyer,
  Maryam Fazel-Zarandi, and Asli Celikyilmaz. 2023.
\newblock \href {https://openreview.net/forum?id=xYlJRpzZtsY} {{ROSCOE}: A
  suite of metrics for scoring step-by-step reasoning}.
\newblock In \emph{International Conference on Learning Representations}.

\bibitem[{Gritsevskiy et~al.(2024)Gritsevskiy, Panickssery, Kirtland, Kauffman,
  Gundlach, Gritsevskaya, Cavanagh, Chiang, La~Roux, and
  Hung}]{gritsevskiy2024rebus}
Andrew Gritsevskiy, Arjun Panickssery, Aaron Kirtland, Derik Kauffman, Hans
  Gundlach, Irina Gritsevskaya, Joe Cavanagh, Jonathan Chiang, Lydia La~Roux,
  and Michelle Hung. 2024.
\newblock \href {https://arxiv.org/abs/2401.05604} {{REBUS}: A robust
  evaluation benchmark of understanding symbols}.
\newblock \emph{Preprint}, arXiv:2401.05604.

\bibitem[{Huang et~al.(2024)Huang, Chen, Mishra, Zheng, Yu, Song, and
  Zhou}]{huang2024selfcorrect}
Jie Huang, Xinyun Chen, Swaroop Mishra, Huaixiu~Steven Zheng, Adams~Wei Yu,
  Xinying Song, and Denny Zhou. 2024.
\newblock Large language models cannot self-correct reasoning yet.
\newblock In \emph{International Conference on Learning Representations}.

\bibitem[{Kasaei et~al.(2026)Kasaei, Marioriyad, Khaleti, Fazli, Baghshah, and
  Rohban}]{kasaei2026rebusbench}
Seyed~Amir Kasaei, Arash Marioriyad, Mahbod Khaleti, MohammadAmin Fazli,
  Mahdieh~Soleymani Baghshah, and Mohammad~Hossein Rohban. 2026.
\newblock \href {https://arxiv.org/abs/2604.01764} {Hidden meanings in plain
  sight: {RebusBench} for evaluating cognitive visual reasoning}.
\newblock \emph{Preprint}, arXiv:2604.01764.

\bibitem[{Kojima et~al.(2022)Kojima, Gu, Reid, Matsuo, and
  Iwasawa}]{kojima2022zeroshot}
Takeshi Kojima, Shixiang Gu, Machel Reid, Yutaka Matsuo, and Yusuke Iwasawa.
  2022.
\newblock Large language models are zero-shot reasoners.
\newblock In \emph{Advances in Neural Information Processing Systems},
  volume~35.

\bibitem[{Kraaijveld et~al.(2025)Kraaijveld, Jiang, Ma, and
  Ilievski}]{kraaijveld2025columbus}
Koen Kraaijveld, Yifan Jiang, Kaixin Ma, and Filip Ilievski. 2025.
\newblock \href {https://doi.org/10.1609/aaai.v39i4.32464} {{COLUMBUS}:
  Evaluating cognitive lateral understanding through multiple-choice rebuses}.
\newblock \emph{Proceedings of the AAAI Conference on Artificial Intelligence},
  39(4):4410--4418.

\bibitem[{Kumar et~al.(2025)Kumar, Zhuang, Agarwal, Su, Co-Reyes, Singh,
  Baumli, Iqbal, Bishop, Roelofs, Zhang, McKinney, Shrivastava, Paduraru,
  Tucker, Precup, Behbahani, and Faust}]{kumar2025score}
Aviral Kumar, Vincent Zhuang, Rishabh Agarwal, Yi~Su, JD~Co-Reyes, Avi Singh,
  Kate Baumli, Shariq Iqbal, Colton Bishop, Rebecca Roelofs, Lei Zhang, Kay
  McKinney, Disha Shrivastava, Cosmin Paduraru, George Tucker, Doina Precup,
  Feryal Behbahani, and Aleksandra Faust. 2025.
\newblock \href
  {https://proceedings.iclr.cc/paper_files/paper/2025/hash/871ac99fdc5282d0301934d23945ebaa-Abstract-Conference.html}
  {Training language models to self-correct via reinforcement learning}.
\newblock In \emph{International Conference on Learning Representations}.

\bibitem[{Lanham et~al.(2023)Lanham, Chen, Radhakrishnan
  et~al.}]{lanham2023faithfulness}
Tamera Lanham, Anna Chen, Ansh Radhakrishnan, et~al. 2023.
\newblock \href {https://arxiv.org/abs/2307.13702} {Measuring faithfulness in
  chain-of-thought reasoning}.
\newblock \emph{Preprint}, arXiv:2307.13702.

\bibitem[{Lee et~al.(2025)Lee, Ge, Wu, Kang, Darrell, and
  Chan}]{lee2025puzzled}
Heekyung Lee, Jiaxin Ge, Tsung-Han Wu, Minwoo Kang, Trevor Darrell, and
  David~M. Chan. 2025.
\newblock \href {https://doi.org/10.18653/v1/2025.emnlp-main.1101} {Puzzled by
  puzzles: When vision-language models can't take a hint}.
\newblock In \emph{Proceedings of the 2025 Conference on Empirical Methods in
  Natural Language Processing}, pages 21724--21737. Association for
  Computational Linguistics.

\bibitem[{Lightman et~al.(2024)Lightman, Kosaraju, Burda, Edwards, Baker, Lee,
  Leike, Schulman, Sutskever, and Cobbe}]{lightman2024verify}
Hunter Lightman, Vineet Kosaraju, Yuri Burda, Harrison Edwards, Bowen Baker,
  Teddy Lee, Jan Leike, John Schulman, Ilya Sutskever, and Karl Cobbe. 2024.
\newblock \href
  {https://proceedings.iclr.cc/paper_files/paper/2024/hash/aca97732e30bcf1303bc22ac3924fd16-Abstract-Conference.html}
  {Let's verify step by step}.
\newblock In \emph{International Conference on Learning Representations}.

\bibitem[{MacGregor and Cunningham(2008)}]{macgregor2008rebus}
James~N. MacGregor and J.~Barton Cunningham. 2008.
\newblock \href {https://doi.org/10.3758/BRM.40.1.263} {Rebus puzzles as
  insight problems}.
\newblock \emph{Behavior Research Methods}, 40(1):263--268.

\bibitem[{Madaan et~al.(2023)Madaan, Tandon, Gupta, Hallinan, Gao, Wiegreffe,
  Alon, Dziri, Prabhumoye, Yang, Gupta, Majumder, Hermann, Welleck,
  Yazdanbakhsh, and Clark}]{madaan2023selfrefine}
Aman Madaan, Niket Tandon, Prakhar Gupta, Skyler Hallinan, Luyu Gao, Sarah
  Wiegreffe, Uri Alon, Nouha Dziri, Shrimai Prabhumoye, Yiming Yang, Shashank
  Gupta, Bodhisattwa~Prasad Majumder, Katherine Hermann, Sean Welleck, Amir
  Yazdanbakhsh, and Peter Clark. 2023.
\newblock \href
  {https://proceedings.neurips.cc/paper_files/paper/2023/hash/91edff07232fb1b55a505a9e9f6c0ff3-Abstract-Conference.html}
  {{Self-Refine}: Iterative refinement with self-feedback}.
\newblock In \emph{Advances in Neural Information Processing Systems},
  volume~36.

\bibitem[{Movva(2025)}]{movva2025reasoning}
Prahitha Movva. 2025.
\newblock \href {https://arxiv.org/abs/2510.02780} {Reasoning riddles: How
  explainability reveals cognitive limits in vision-language models}.
\newblock \emph{Preprint}, arXiv:2510.02780.

\bibitem[{Prasad et~al.(2023)Prasad, Saha, Zhou, and
  Bansal}]{prasad2023receval}
Archiki Prasad, Swarnadeep Saha, Xiang Zhou, and Mohit Bansal. 2023.
\newblock \href {https://doi.org/10.18653/v1/2023.emnlp-main.622} {{ReCEval}:
  Evaluating reasoning chains via correctness and informativeness}.
\newblock In \emph{Proceedings of the 2023 Conference on Empirical Methods in
  Natural Language Processing}, pages 10066--10086. Association for
  Computational Linguistics.

\bibitem[{Song et~al.(2025)Song, Su, Qu, Zhou, and Cheng}]{song2025prmbench}
Mingyang Song, Zhaochen Su, Xiaoye Qu, Jiawei Zhou, and Yu~Cheng. 2025.
\newblock \href {https://doi.org/10.18653/v1/2025.acl-long.1230} {{PRMBench}: A
  fine-grained and challenging benchmark for process-level reward models}.
\newblock In \emph{Proceedings of the 63rd Annual Meeting of the Association
  for Computational Linguistics (Volume 1: Long Papers)}, pages 25299--25346.
  Association for Computational Linguistics.

\bibitem[{Turpin et~al.(2023)Turpin, Michael, Perez, and
  Bowman}]{turpin2023unfaithful}
Miles Turpin, Julian Michael, Ethan Perez, and Samuel~R. Bowman. 2023.
\newblock Language models don't always say what they think: Unfaithful
  explanations in chain-of-thought prompting.
\newblock In \emph{Advances in Neural Information Processing Systems},
  volume~36.

\bibitem[{Tyen et~al.(2024)Tyen, Mansoor, Carbune, Chen, and
  Mak}]{tyen2024errorlocation}
Gladys Tyen, Hassan Mansoor, Victor Carbune, Peter Chen, and Tony Mak. 2024.
\newblock \href {https://doi.org/10.18653/v1/2024.findings-acl.826} {{LLMs}
  cannot find reasoning errors, but can correct them given the error location}.
\newblock In \emph{Findings of the Association for Computational Linguistics:
  ACL 2024}, pages 13894--13908. Association for Computational Linguistics.

\bibitem[{Wei et~al.(2022)Wei, Wang, Schuurmans, Bosma, Ichter, Xia, Chi, Le,
  and Zhou}]{wei2022cot}
Jason Wei, Xuezhi Wang, Dale Schuurmans, Maarten Bosma, Brian Ichter, Fei Xia,
  Ed~H. Chi, Quoc~V. Le, and Denny Zhou. 2022.
\newblock Chain-of-thought prompting elicits reasoning in large language
  models.
\newblock In \emph{Advances in Neural Information Processing Systems},
  volume~35.

\bibitem[{Xia et~al.(2025)Xia, Li, Liu, Wu, and Liu}]{xia2025reasoneval}
Shijie Xia, Xuefeng Li, Yixin Liu, Tongshuang Wu, and Pengfei Liu. 2025.
\newblock \href {https://doi.org/10.1609/aaai.v39i26.34987} {Evaluating
  mathematical reasoning beyond accuracy}.
\newblock \emph{Proceedings of the AAAI Conference on Artificial Intelligence},
  39(26):27723--27730.

\bibitem[{Yang et~al.(2025{\natexlab{a}})Yang, Wu, Chen, Xiao, Yang, Wong, and
  Wang}]{yang2025ahamoments}
Shu Yang, Junchao Wu, Xin Chen, Yunze Xiao, Xinyi Yang, Derek~F. Wong, and
  Di~Wang. 2025{\natexlab{a}}.
\newblock \href {https://arxiv.org/abs/2504.02956} {Understanding {Aha}
  moments: from external observations to internal mechanisms}.
\newblock \emph{Preprint}, arXiv:2504.02956.

\bibitem[{Yang et~al.(2025{\natexlab{b}})Yang, Zhang, Wang, Xu, Lin, and
  Sui}]{yang2025confidence}
Zhe Yang, Yichang Zhang, Yudong Wang, Ziyao Xu, Junyang Lin, and Zhifang Sui.
  2025{\natexlab{b}}.
\newblock \href {https://doi.org/10.18653/v1/2025.acl-long.203} {Confidence
  v.s. critique: A decomposition of self-correction capability for {LLMs}}.
\newblock In \emph{Proceedings of the 63rd Annual Meeting of the Association
  for Computational Linguistics (Volume 1: Long Papers)}, pages 3998--4014.
  Association for Computational Linguistics.

\bibitem[{Yao et~al.(2023)Yao, Yu, Zhao, Shafran, Griffiths, Cao, and
  Narasimhan}]{yao2023tot}
Shunyu Yao, Dian Yu, Jeffrey Zhao, Izhak Shafran, Thomas~L. Griffiths, Yuan
  Cao, and Karthik Narasimhan. 2023.
\newblock Tree of thoughts: Deliberate problem solving with large language
  models.
\newblock In \emph{Advances in Neural Information Processing Systems},
  volume~36.

\bibitem[{Yu et~al.(2025)Yu, Li, and Wang}]{yu2025scalingflaws}
Fei Yu, Yingru Li, and Benyou Wang. 2025.
\newblock \href {https://arxiv.org/abs/2502.00271} {Scaling flaws of
  verifier-guided search in mathematical reasoning}.
\newblock \emph{Preprint}, arXiv:2502.00271.

\bibitem[{Zhang et~al.(2025)Zhang, Feng, Ni, Cao, Liu, Avestimehr, Butler,
  Weng, Zhang, Narayanan, and Avestimehr}]{zhang2025chinese}
Tuo Zhang, Tiantian Feng, Yibin Ni, Mengqin Cao, Ruying Liu, Kiana Avestimehr,
  Katharine Butler, Yanjun Weng, Mi~Zhang, Shrikanth Narayanan, and Salman
  Avestimehr. 2025.
\newblock \href {https://doi.org/10.18653/v1/2025.findings-acl.1155} {Creating
  a lens of {Chinese} culture: A multimodal dataset for {Chinese} pun rebus art
  understanding}.
\newblock In \emph{Findings of the Association for Computational Linguistics:
  ACL 2025}, pages 22473--22487. Association for Computational Linguistics.

\bibitem[{Zhao et~al.(2025)Zhao, Sun, Shi, and Song}]{zhao2025fakeaha}
Jiachen Zhao, Yiyou Sun, Weiyan Shi, and Dawn Song. 2025.
\newblock Can aha moments be fake? identifying true and decorative thinking
  steps in chain-of-thought.
\newblock \emph{arXiv preprint arXiv:2510.24941}.

\bibitem[{Zheng et~al.(2025)Zheng, Zhang, Zhang, Lin, Lu, Yu, Liu, Zhou, and
  Lin}]{zheng2025processbench}
Chujie Zheng, Zhenru Zhang, Beichen Zhang, Runji Lin, Keming Lu, Bowen Yu,
  Dayiheng Liu, Jingren Zhou, and Junyang Lin. 2025.
\newblock \href {https://doi.org/10.18653/v1/2025.acl-long.50} {{ProcessBench}:
  Identifying process errors in mathematical reasoning}.
\newblock In \emph{Proceedings of the 63rd Annual Meeting of the Association
  for Computational Linguistics (Volume 1: Long Papers)}, pages 1009--1024.
  Association for Computational Linguistics.

\end{thebibliography}


\clearpage
\appendix
\addtocontents{toc}{\protect\setcounter{tocdepth}{2}}
\begingroup
\hypersetup{linkcolor=black}
\renewcommand{\contentsname}{Appendix Contents}
\tableofcontents
\endgroup
\clearpage

\section{Additional Benchmark Details}
\label{app:benchmark-details}

\subsection{CoT Annotation Protocol}
\label{app:annotation-protocol}

Table~\ref{tab:annotation-protocol} summarizes the annotation workflow
used for visible reasoning traces. Candidate detection is recall-oriented:
explicit reconsideration phrases are recorded first. Multiple trigger phrases along the same continuous exploration path are grouped as one Aha candidate. Aha candidates with no
qualifying substantive state change remain in the candidate-only records
and contribute to candidate counts, while substantive metrics use the
events that satisfy the subsequent judgment criteria.

\paragraph{Evidence representation.}
The evidence component of $H_t$ is decomposed as
\begin{equation}
E_t = (\mathcal O_t,\mathcal M_t),
\label{eq:evidence-state}
\end{equation}
where $\mathcal O_t$ denotes the clues already identified and used,
and $\mathcal M_t$ denotes the interpretation mechanisms currently
employed.

\begin{table*}[!tp]
\centering
\small
\setlength{\tabcolsep}{5pt}
\renewcommand{\arraystretch}{1.12}

\begin{tabular}{p{0.14\textwidth} p{0.26\textwidth} p{0.51\textwidth}}
\toprule
\textbf{Stage} & \textbf{Annotation target} & \textbf{Operational rule} \\
\midrule
Segmentation
& Semantic reasoning units
& Introduce a boundary when the reasoning function changes, including
hypothesis formation or revision, clue use, verification, conflict
detection, mechanism switching, or candidate adoption/rejection. \\

State reconstruction
& $H_t=(\mathcal A_t,E_t,V_t)$
& Record the current candidates, used evidence, active interpretation
mechanisms, and candidate--evidence relations after each unit. \\

Candidate detection
& Aha candidate
& Mark an Aha candidate when reflective language, failed
verification, conflict, candidate switching, mechanism switching, or
new evidence use is observed. \\

Substantive judgment
& Substantive Aha
& Require a prior state, an identifiable trigger, a core state
revision, and explicit evidence grounding. \\

Outcome labeling
& Aha outcome
& Record whether an explored revision is retained in subsequent reasoning,
and compare the retained or best explored state with the pre-event state
according to the outcome taxonomy. \\
\bottomrule
\end{tabular}
\caption{Process-level annotation protocol for visible reasoning traces.}
\label{tab:annotation-protocol}

\end{table*}

\begin{figure*}[p]
\centering
\begin{minipage}{\textwidth}
\fontsize{10}{12}\selectfont
\setlength{\parindent}{0pt}
\setlength{\parskip}{4pt}
\textbf{Worked Annotation Example}\par
\begin{minipage}[c]{0.23\linewidth}
\centering
\includegraphics[width=0.95\linewidth]{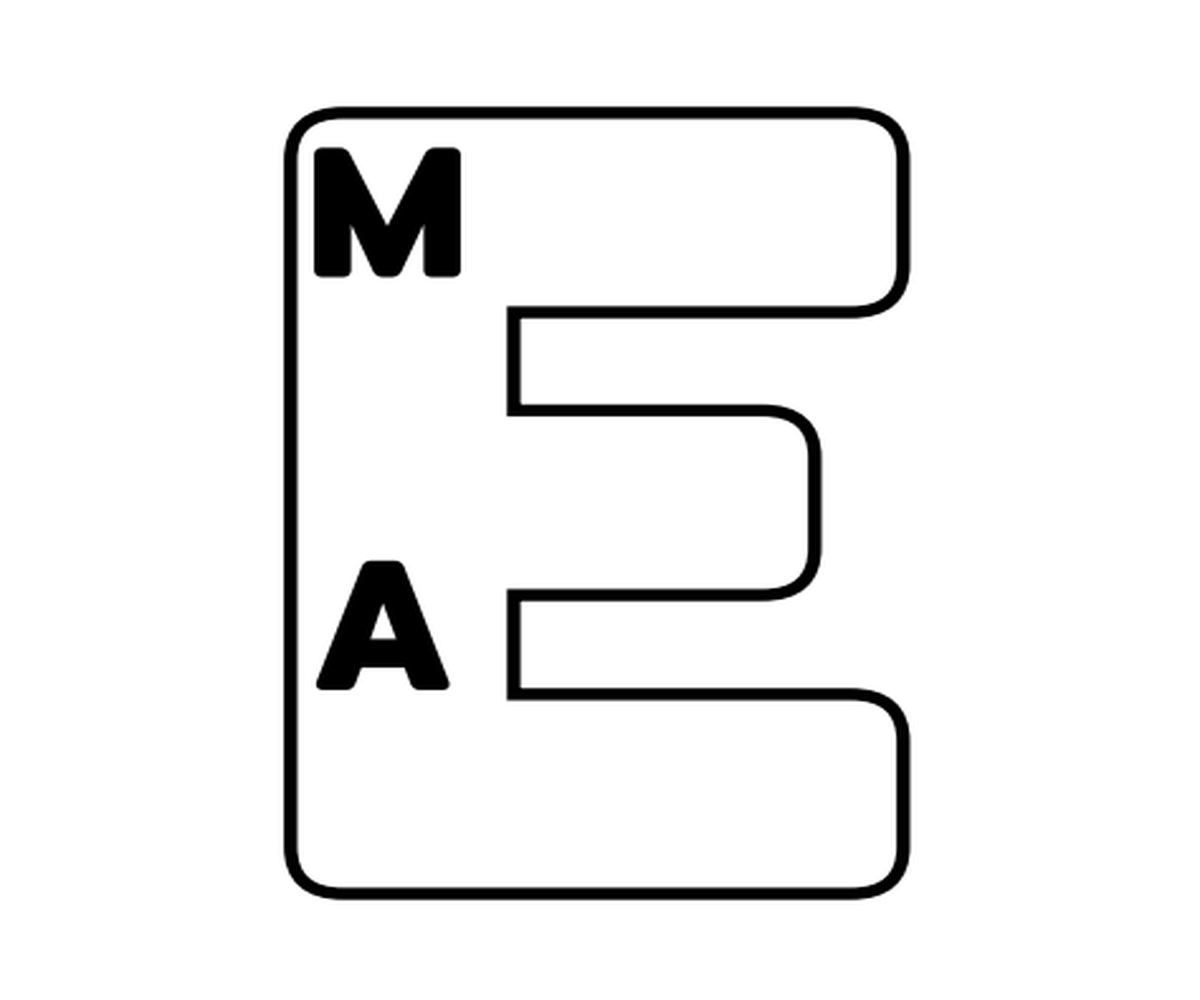}
\end{minipage}\hfill
\begin{minipage}[c]{0.73\linewidth}
\textbf{Task:} English rebus puzzle\par
\textbf{Model:} Gemini 3.5 Flash\par
\textbf{Gold answer:} Maine\par
\end{minipage}
\par\medskip\hrule\medskip
\begin{minipage}[t]{0.48\linewidth}
\raggedright
\textbf{CoT units}\par
\textbf{$u_1$.} The image shows a large letter E, with M and A inside its left-hand region (or within its outline). M and A are inside E, so this can be interpreted as ``M and A in E.''\par
\textbf{$u_2$.} Reading ``M and A in E'' together sounds very similar to ``man-in-the-moon'' or ``men in...''. \textbf{No, let us sound it out carefully:} ``M'' (em), ``A'' (ay), ``in'', ``E'' (ee) $\rightarrow$ ``M A in E'' $\rightarrow$ ``Maine''?\par
\textbf{$u_3$.} \textbf{Let us reconsider:} ``M'' and ``A'' in ``E'' $\rightarrow$ ``M-A-in-E'' $\rightarrow$ ``Maine.''\par
\textbf{$u_4$.} This is a very classic English rebus puzzle. The letters M and A are inside E, giving M-A-in-E, which spells the US state Maine.
\end{minipage}\hfill
\begin{minipage}[t]{0.48\linewidth}
\raggedright
\textbf{Candidate detection and substantive judgment}\par
\textbf{$u_2$: substantive Aha; Functional.}
``No, let us sound it out carefully'' triggers reconsideration. The model revises its tentative candidates and interpretation mechanism, using the spatial relation to construct Maine. Subsequent units retain this answer and explanation.\par
The pre-revision state is taken from the text preceding ``No'' within $u_2$. The revision satisfies all four substantive criteria: a prior state, an identifiable trigger, a core revision, and evidence grounding. Its adoption is supported by $u_3$, $u_4$, and the final answer.\par
\textbf{$u_3$: candidate-only.}
``Let us reconsider'' repeats Maine and the same interpretation. No further core revision occurs in $A$, $E$, or $V$.
\end{minipage}
\par\medskip
\textbf{State revision within $u_2$}\par
\setlength{\tabcolsep}{4pt}
\renewcommand{\arraystretch}{1.08}
\begin{tabularx}{\linewidth}{@{}p{0.13\linewidth}>{\raggedright\arraybackslash}X>{\raggedright\arraybackslash}X@{}}
\toprule
\textbf{Component} & \textbf{Before the trigger} & \textbf{After revision} \\
\midrule
$A$ & Tentative candidates ``man-in-the-moon'' and ``men in...'' & Maine \\
$E$ & The identified letters and containment relation are used for a loose phonetic association. & The containment relation supplies IN, yielding M + A + IN + E = MAINE. \\
$V$ & The tentative candidates have a weak correspondence with the letters and their arrangement. & The revised candidate accounts for both the letters and their containment relation. \\
\bottomrule
\end{tabularx}
\par\medskip
\textbf{State-quality scores}\par
\begin{tabularx}{\linewidth}{@{}>{\raggedright\arraybackslash}X>{\raggedright\arraybackslash}p{0.27\linewidth}rrrr@{}}
\toprule
\textbf{State} & \textbf{Candidate} & $q_A$ & $q_E$ & $q_V$ & $Q_{\mathrm{total}}$ \\
\midrule
Before revision, within $u_2$ & man-in-the-moon / men in... & 0 & 0.5 & 0.5 & 0.333 \\
Best explored state, within $u_2$ & Maine & 1 & 1 & 1 & 1.000 \\
Adopted state, confirmed in $u_3$--$u_4$ & Maine & 1 & 1 & 1 & 1.000 \\
\bottomrule
\end{tabularx}
\par
{\footnotesize The scores reproduce the existing annotation; $q_E$ remains provisional.}\par
\textbf{Scoring rationale.}
Before revision, the tentative answers are incorrect ($q_A=0$). The letters and containment relation have been identified, but their interpretation remains incomplete ($q_E=0.5$). The tentative answers only partially correspond to the evidence ($q_V=0.5$). After revision, Maine is correct and follows from M + A + IN + E. This interpretation accounts for the letters and their spatial relation, giving $(q_A,q_E,q_V)=(1,1,1)$.
\[
\Delta Q_{\mathrm{total}}
=\frac{(1-0)+(1-0.5)+(1-0.5)}{3}
=\frac{2}{3}\approx 0.667.
\]
The improved state is retained in subsequent reasoning, so the event is classified as Functional Aha.
\end{minipage}
\caption{Worked annotation example for the English rebus puzzle Maine.}
\label{fig:maine-annotation-example}
\end{figure*}

\paragraph{Worked annotation example.}
Figure~\ref{fig:maine-annotation-example} presents a complete example with
candidate judgments, state revisions, and state-quality scores.

\subsection{State-quality Definition}
\label{app:state-quality}

For puzzle $i$, let
\begin{equation}
G_i=
(\mathcal O_i^{*},\mathcal M_i^{*},\mathcal R_i^{*})
\label{eq:gold-evidence}
\end{equation}
denote its gold evidence, where $\mathcal O_i^{*}$ contains the required
clues, $\mathcal M_i^{*}$ the required interpretation mechanisms, and
$\mathcal R_i^{*}$ their relations to the gold answer.

The answer component is
\begin{equation}
q_A(H)
=
\mathbb I
\left[
a^{\mathrm{pref}}(H)=y_i
\right].
\label{eq:answer-quality}
\end{equation}

Evidence coverage $q_E(H)$ uses five ordinal anchors: $0$ for largely incorrect or missing key clues and mechanisms, $0.25$ for a few relevant surface clues with an unstable mechanism, $0.5$ for partial recovery of key clues or mechanisms, $0.75$ for substantial recovery with important omissions, and $1$ for agreement with the reference evidence and mechanisms.

Candidate--evidence consistency $q_V(H)$ uses three anchors: $0$ for conflict or lack of support from the main evidence, $0.5$ for partial support with important omissions or strained explanations, and $1$ when the main clues support the current candidate. These scores follow the operational annotation rubric.

We use equal weights across the three dimensions:
\begin{equation}
Q_{\mathrm{total}}(H)
=
\frac{
q_A(H)+q_E(H)+q_V(H)
}{3}.
\label{eq:total-state-quality}
\end{equation}

\subsection{Aha Outcome Definitions}
\label{app:aha-outcomes}

The quality change used for event classification follows
Eq.~\ref{eq:quality-change}. NQGD compares the actual post-event
and pre-event states. Missed events may still show positive quality
changes in the actual post-event state.

\begin{table*}[!tp]
\centering
\small
\setlength{\tabcolsep}{6pt}
\renewcommand{\arraystretch}{1.12}

\begin{tabular}{lccc p{0.39\textwidth}}
\toprule
\textbf{Outcome}
& \textbf{Better state}
& \textbf{Adopted}
& $\boldsymbol{\Delta Q_{\mathrm{total}}}$
& \textbf{Interpretation} \\
\midrule
Missed
& Yes & No & $>0$
& A better state is discovered during exploration but not retained. \\

Unresolved
& No & No & $\leq0$
& Reconsideration occurs without producing a better state. \\

Functional
& Yes & Yes & $>0$
& The improved state is adopted and yields a positive quality gain. \\

Ineffective
& -- & Yes & $=0$
& A revised state is adopted without net improvement. \\

Regressive
& -- & Yes & $<0$
& The adopted revision decreases state quality. \\
\bottomrule
\end{tabular}
\caption{Outcome taxonomy for substantive Aha events.}
\label{tab:aha-outcomes}

\end{table*}

\subsection{Metric Definitions}
\label{app:metric-definitions}

Let $T$ denote the total number of characters in visible reasoning
traces. For Aha candidate $k$, define
\begin{equation}
\begin{aligned}
C_k&=\mathbb I[r_k\lor x_k\lor b_k\lor m_k\lor n_k],\\
S_k&=C_kp_kt_ku_kg_k.
\end{aligned}
\label{eq:candidate-substantive-criteria}
\end{equation}

For aggregation, let $K=\sum_k C_k$ count Aha candidates and let
$\mathcal E$ index the unique substantive events, with $S=|\mathcal E|$.
Candidate--event links need not be one-to-one; each substantive event
is counted once. Let $F$ count Functional events and $B$ count Functional
plus Missed events, following the recorded outcome labels.

NQGD measures local changes at annotated event endpoints. Their sum
need not equal the quality difference between the initial and final
states of a CoT.

\begin{table*}[!tp]
\centering
\small
\setlength{\tabcolsep}{5pt}
\renewcommand{\arraystretch}{1.18}

\begin{tabular}{l p{0.31\textwidth} p{0.48\textwidth}}
\toprule
\textbf{Metric} & \textbf{Definition} & \textbf{Interpretation} \\
\midrule

$\mathrm{CED}_{10k}$
&
$\displaystyle
\frac{10^4K}{T}$
&
Density of Aha candidates in visible reasoning. \\

$\mathrm{SY}$
&
$\displaystyle
\frac{S}{K}$
&
Corpus-level substantive-event yield per Aha candidate. \\

$\mathrm{BDR}$
&
$\displaystyle
\frac{B}{S}$
&
Fraction of substantive revisions that discover a better state. \\

$\mathrm{DAR}$
&
$\displaystyle
\frac{F}{B}$
&
Fraction of discovered improvements converted into Functional Aha. \\

$\mathrm{FED}_{10k}$
&
$\displaystyle
\frac{10^4F}{T}$
&
Density of Functional Aha events. \\

$\mathrm{NQGD}_{10k}$
&
$\displaystyle
\frac{10^4}{T}\sum_{k\in\mathcal E}
\boldsymbol{\lambda}^{\top}\left[Q(H_{k+1})-Q(H_k)\right]$
&
Observed net quality change at substantive Aha event endpoints per
10,000 CoT characters. \\

\midrule

$\mathrm{CDR}$
&
$\displaystyle
\frac{1}{N}\sum_{i=1}^{N} d_i$
&
Fraction of samples in which the model explicitly detects a conflict. \\

$\mathrm{CRR}$
&
$\displaystyle
\frac{\sum_i d_i f_i}{\sum_i d_i}$
&
Fraction of conflict-containing samples with a subsequent Functional Aha. \\

$\mathrm{FR}$
&
$\displaystyle
\frac{\sum_i d_i r_i(1-f_i)}{\sum_i d_i}$
&
Fraction of conflict cases that retain the previous path without a
functional correction. \\

\bottomrule
\end{tabular}
\caption{Formal definitions of the process-level diagnostic metrics.}
\label{tab:metric-definitions}

\end{table*}

Here, $d_i$ indicates that sample $i$ contains an explicitly detected
conflict, $r_i$ indicates that the pre-conflict preferred candidate is
retained, and $f_i$ indicates that a Functional Aha occurs after the
detected conflict. An undefined metric is reported as -- when its denominator is zero.

\subsection{Puzzle Mechanism Taxonomy}
\label{app:puzzle_taxonomy}

Table~\ref{tab:puzzle_taxonomy} defines the categories used in
Section~\ref{sec:language_category_analysis} and gives puzzle counts
for the 300 Chinese and 100 English evaluation items.
Labels are non-exclusive: recognition requirements and multiple
reasoning mechanisms may co-occur, so column totals exceed the
number of puzzles. Counts refer to unique puzzles, independent of
the number of evaluated models.

English labels retain the author-provided skill annotations from
\emph{Puzzled by Puzzles}~\citep{lee2025puzzled}.
Chinese labels were assigned through model-assisted review of
the puzzle images and gold solutions using the corresponding
category definitions. Chinese character and word manipulation
(CWM) is aligned with English letter and word manipulation (LWM)
at the broad category level; language-specific operations are
preserved. The counts use the current labels, including 11
provisional Chinese assignments recorded in the annotation files.

\begin{table*}[!tp]
\centering
\small
\setlength{\tabcolsep}{4pt}
\renewcommand{\arraystretch}{1.12}

\begin{tabularx}{\textwidth}{@{}l>{\raggedright\arraybackslash}p{0.22\textwidth}>{\raggedright\arraybackslash}Xrr@{}}
\toprule
Code & Category & Operational definition & Chinese & English \\
 & & & ($n=300$) & ($n=100$) \\
\midrule
IR & Image recognition & Identify objects, people, actions, or depicted symbols. & 258 & 79 \\
TR & Text recognition & Read written words or characters, including stylized text. & 219 & 74 \\
PW & Phonetics and wordplay & Use homophones, similar sounds, or puns to derive the answer. & 115 & 20 \\
SPR & Spatial and positional reasoning & Use relations such as above/below, inside/outside, or adjacency. & 85 & 19 \\
FS & Font style, size, and color & Interpret differences in typeface, size, or color as clues. & 40 & 12 \\
CWM/LWM & Character, letter, and word manipulation & Interpret components, overlap, concealment, repetition, or rearrangement of text. & 27 & 11 \\
SS & Symbolic substitution & Replace a number, letter, or pictorial symbol with a word or meaning. & 37 & 4 \\
VMCR & Visual metaphors and cultural references & Connect visual clues to metaphors, idioms, or cultural references. & 31 & 7 \\
QMR & Quantitative or mathematical reasoning & Use counts, mathematical symbols, or quantitative relations. & 29 & 1 \\
AN & Absence or negation & Interpret missing, removed, or crossed-out elements, or a depicted negation. & 21 & 3 \\
TO & Text orientation & Use rotation, inversion, or another change in text direction. & 8 & 2 \\
\bottomrule
\end{tabularx}
\caption{Puzzle categories, definitions, and counts in the evaluation set.
CWM and LWM denote the Chinese and English forms of text manipulation,
respectively. Each puzzle can carry multiple labels.}
\label{tab:puzzle_taxonomy}

\end{table*}

\section{Experimental Details and Provenance}
\label{app:experimental-details}

\subsection{Evaluation Corpora}
Full-corpus accuracy counts missing
final answers as incorrect. The evaluation set contains 300 Chinese and 100 English puzzles. The Chinese extension adds 100 puzzles (30 idiom, 23 food, 46 other visual-rebus puzzles, and one phrase) to the archived 200. Each model is evaluated on the same 400 assigned puzzles. Missing responses are counted as failures for full-set accuracy. Process rates use observable traces; event rates use their corresponding candidate or event denominators. Densities are normalized by recorded CoT characters.The cross-task pilot includes ten models, with 12 problems assigned per task. Excluding events labeled uncertain, Rebus, AIME2024, and MATH500 yield 63, 38, and 3 substantive Aha events across 120, 119, and 120 observable CoT traces, respectively.

\subsection{Inference and Annotation Records}
The stored configuration records identify GPT-5.5 and GPT-5.6 Sol with
reasoning effort \texttt{none}, MiniMax M3 with thinking disabled, and
Opus 4.8 with provider-side thinking disabled. Gemini 3.5 Flash records
\texttt{minimal} reasoning effort and no returned provider-reasoning text;
this record does not verify that provider-side thinking was fully disabled.
Model display names identify the evaluated service aliases; provider
aliases are not immutable weight snapshots. The comparison combines
archived runs and does not imply a simultaneous, identical-budget run of
all models. The added runs retain the archived task prompts, reasoning settings, and output limits. GPT-5.6 Sol and Opus 4.8 omit temperature because their current endpoints reject the archived value.

Aha labels use model-assisted item-level annotation followed by stored
validation and targeted semantic review. Initial-preferred-answer labels
were extracted separately from trace-only packets, with external gold,
final-answer fields, and pre-existing Aha quality labels withheld during
extraction. The trace's own terminal answer remains visible in its text.
The extraction records include exact supporting quotations and offsets.
Labels were frozen before aggregate answer-transition analysis. Central
review includes all-record key/hash/quote validation and targeted semantic
audits; it is not independent human double annotation of every trace.
Initial-answer labels are available for all stored main-evaluation responses. The main process
summaries and the gap analysis use existing event annotations without
relabeling them from answer-transition outcomes.

\paragraph{Cross-model annotation agreement.}
We assessed annotation agreement between Qwen3.8 Max and GPT-5.5 on 50 CoT traces sampled across generating models, trace lengths, and annotation categories. Both annotators followed the same protocol, distinguishing Aha candidates from substantive events. Events were matched one-to-one using unit-span IoU $\geq 0.25$. We report presence agreement for substantive Aha, Functional Aha, and conflicts, together with event-level F1.

\subsection{Information-assistance Subset}
The three information conditions share 100 puzzle IDs for each model:
45 Zhihu-source items and 55 traditional-culture-source items selected
from the baseline pool. Five records with gold-quality flags remain in
the frozen set. T0 reuses historical outputs; T1 and T2 are subsequent
assistance runs. Thus, the pairings control puzzle identity while retaining
run-time and service-alias differences. Facts and mechanism hints were
prepared and reviewed with model assistance. T2 was designed after the
Opus T1 pilot and frozen before the new Qwen answers.

The Qwen assistance configurations use temperature 0, maximum output
16,000 tokens, thinking disabled, and no system prompt. The Opus
assistance configuration uses OpenRouter, temperature 0, maximum output
16,000 tokens, and reasoning effort \texttt{none}. Its system message is
reported in Appendix~\ref{app:prompts}. Historical T0 annotations, the
Opus T1 pilot, and the later T2 review retain different annotation
provenance. Appendix~\ref{app:information-details} reports raw event
denominators and the available boundary sensitivity.

\section{Task and Assistance Prompts}
\label{app:prompts}

The assistance conditions preserve the image and the base task template.
The following example is an English translation of the stored prompt preview for
sample \texttt{zhihu\_511815257\_011}. The fields request visible reasoning,
a final answer, confidence, and a brief explanation. Observation and
mechanism additions do not insert the gold answer. The word ``verified''
is part of the recorded prompt header; the preparation procedure is the
model-assisted review described in Appendix~\ref{app:experimental-details}.
\par\medskip
\begin{ahaprompt}{Base user prompt}
\prompttag{T0, T1, T2}\hspace{5pt}{\footnotesize User message}\par\medskip
\promptheading{TASK}
Think through this Chinese visual rebus puzzle step by step, then provide the answer.\par
Sample ID: zhihu\_511815257\_011\par
Puzzle category: chengyu\par
\tcblower
\promptheading{OUTPUT FORMAT}
Output only one JSON object in your final response. Do not use Markdown:\par
\texttt{\{}\par
\texttt{"visible\_cot":} ``Complete step-by-step reasoning'',\par
\texttt{"final\_answer":} ``Final answer'',\par
\texttt{"confidence": 0.0,}\par
\texttt{"brief\_reason":} ``A one-sentence explanation of the key clue''\par
\texttt{\}}
\end{ahaprompt}

\begin{ahaprompt}{Observation addition}
\prompttag{T1, T2}\hspace{5pt}{\footnotesize Added to the user prompt}\par\medskip
[Additional verified observations]\par
The image contains a large black Chinese character ``步''. A yellow star shape appears at its lower right, with several long curved lines trailing behind it.
\end{ahaprompt}
\par\medskip
\begin{ahaprompt}{Mechanism addition}
\prompttag{T2}\hspace{5pt}{\footnotesize Added to the user prompt}\par\medskip
[Additional reasoning mechanism hint]\par
You may prioritize the following solution mechanisms:\par\smallskip
- Literal morpheme composition: Map text, objects, and actions to corresponding morphemes, combine them into a common Chinese idiom, and check the main clues.\par\smallskip
- Quantity and scale encoding: Use relationships involving the number, size, and length of elements to construct an idiom, while retaining other textual or object clues.
\end{ahaprompt}

\begin{ahaprompt}{Opus assistance system message}
\prompttag{Opus}\hspace{5pt}{\footnotesize System message}\par\medskip
\promptheading{ROLE}
You are a careful multimodal reasoning model for Chinese visual rebus
puzzles. Follow the user's requested output schema exactly.
\end{ahaprompt}

\section{Additional Results and Statistical Details}
\label{app:additional-results}

\subsection{Answer-layer Aha Transitions}
\label{app:answer-transitions}
Prior work distinguishes preserving correct answers from correcting incorrect ones when evaluating self-correction \cite{yang2025confidence}.The initial answer is the first explicitly preferred candidate before
terminal selection that is followed by substantive checking or reasoning.
Let $C$ and $W$ denote correct and wrong answers under the frozen answer
scoring rule. The first letter of $CC,CW,WC,WW$ refers to the initial
preferred answer and the second to the final answer. Within each paired
subset, $n=CC+CW+WC+WW$, and
\begin{equation}
\begin{aligned}
I&=(CC+CW)/n,\\
F&=(CC+WC)/n,\\
R&=WC/(WC+WW),\\
D&=CW/(CC+CW),\\
\Delta&=100(WC-CW)/n.
\end{aligned}
\label{eq:answer-transition-rates}
\end{equation}
Thus $F=I+(1-I)R-ID$ whenever the conditional denominators are nonzero.
For Aha-LCR, the numerator is the number of $WC$ puzzles containing at
least one Functional Aha with $q_A:0\!\rightarrow\!1$; the denominator is
$WC+WW$. A local Aha gain may have a different temporal boundary from
the first preferred answer and is not proof of causation.

An explicitly preferred candidate must be advanced as the current leading
solution, including a tentative leading hypothesis with substantive
evaluation. Unranked alternatives, OCR strings, and immediately rejected
questions do not qualify. Terminal-only candidates, ambiguous cases,
readable traces without a preferred candidate, and unavailable traces are
recorded separately. Paired analysis additionally requires an available
final answer.

\begin{table*}[t]
\centering
\small
\setlength{\tabcolsep}{4pt}
\renewcommand{\arraystretch}{1.08}

\begin{tabular*}{\textwidth}{@{\extracolsep{\fill}}lrrrrrcr@{}}
\toprule
Model & CC & CW & WC & WW & Aha-linked & $\Delta$ 95\% CI & $p$ \\
\midrule
GPT-5.5 & 6 & 0 & 4 & 64 & 2/68 & [1.4, 10.8] & 0.1250 \\
GPT-5.6 Sol & 2 & 0 & 5 & 37 & 3/42 & [2.3, 20.5] & 0.0625 \\
Gemini 2.5 Flash & 52 & 0 & 8 & 250 & 8/258 & [1.0, 4.5] & 0.0078 \\
Gemini 3.5 Flash & 70 & 3 & 4 & 96 & 3/100 & [-2.3, 3.5] & 1.0000 \\
Claude Opus 4.8 & 35 & 2 & 14 & 202 & 12/216 & [2.0, 7.9] & 0.0042 \\
Kimi K2.6 & 66 & 4 & 7 & 163 & 6/170 & [-1.2, 3.8] & 0.5488 \\
Kimi K3 & 160 & 9 & 18 & 135 & 18/153 & [-0.3, 6.2] & 0.1221 \\
Qwen 3.7 Plus & 90 & 4 & 20 & 183 & 19/203 & [2.4, 8.8] & 0.0015 \\
Qwen 3.8 Max & 95 & 6 & 23 & 78 & 22/101 & [3.5, 13.9] & 0.0023 \\
MiniMax M3 & 24 & 3 & 9 & 221 & 8/230 & [-0.4, 5.1] & 0.1460 \\
Doubao 2.1 Pro & 111 & 2 & 3 & 16 & 2/19 & [-2.3, 4.5] & 1.0000 \\
\bottomrule
\end{tabular*}
\caption{Answer-transition counts and uncertainty. Aha-linked counts give the Aha-LCR numerator/denominator. The net-gain interval is in percentage points.}
\label{tab:answer-counts}
\par\smallskip
\begin{minipage}{\textwidth}\footnotesize\raggedright
Net-gain intervals resample paired puzzles. $p$: exact two-sided McNemar test, unadjusted for multiple comparisons. No significance stars are assigned.
\end{minipage}
\end{table*}

All intervals describe sampling variability within the observed paired subsets and do not include annotation uncertainty. Doubao has 216 pre-final labels and 132 paired endpoints; 40 of its 84 missing endpoints already have a correct initial answer.

\paragraph{Model-level observations.}
Qwen 3.8 Max corrects 23 of 101 initially wrong answers and loses 6 of 101 initially correct answers, yielding an 8.4-point net gain. Qwen 3.7 Plus corrects 20 of 203 errors and loses 4 initially correct answers, for a 5.4-point gain. Opus has 253 paired observations, with 14 corrections and 2 degradations.

GPT-5.5 and GPT-5.6 Sol have 74 and 44 paired observations. Doubao has 113 initially correct answers among 132 paired observations, leaving 19 correction opportunities. Its paired final accuracy is 86.4\%, compared with 41.2\% over the full set, reflecting endpoint selection.

Functional Aha occurs in 171/250 and 114/221 persistently incorrect cases for Gemini 2.5 Flash and MiniMax M3. Local gains in evidence use or consistency can coexist with an incorrect final answer.

\paragraph{Answer-form and common-item checks.}
\label{app:answer-sensitivity}

Removing Gemini 3.5 Flash's closely related-form correction (遮风避雨 to 遮风挡雨) leaves 172 paired observations, 3 corrections among 99 initially wrong answers, and a 0.0-point net gain. The main table retains the frozen scoring rule. The Qwen models share 161 paired puzzles. Initial accuracy is 31.7\% and 50.3\%, and paired final accuracy is 36.6\% and 55.9\% for Qwen 3.7 Plus and Qwen 3.8 Max, respectively. Among 68 puzzles both initially answer incorrectly, correction counts are 4/68 and 10/68. This common subset is selected by observable preferred answers and endpoint availability.

\subsection{Checkpoint Selection and Discovery--Retention Gaps}
\label{app:checkpoint-retention}
For each model--puzzle record, we select the last substantive-Aha event
with complete scored before, explored-best, and after states and a
parseable span. The retained state is the actual after state at that
checkpoint. The best state is the highest-quality complete state recorded
up to and including the checkpoint, considering before states,
declared explored-best states, after states, and the recorded exploration
path. We maximize the quality of complete states; components from
different states are not combined. With this selection,
\begin{equation}
\begin{aligned}
g_i&=\max_{H\in\mathcal H_{i,\leq k_i}}Q_{\mathrm{total}}(H)\\
&\quad-Q_{\mathrm{total}}(H^+_{i,k_i}).
\end{aligned}
\label{eq:discovery-retention-gap}
\end{equation}
All included puzzles, including those with $g_i=0$, enter the mean and
positive-gap denominator. Table~\ref{tab:discovery-retention-gap} reports
model-specific coverage. Across 4,400 assigned model--puzzle evaluations, 2,357 are included; 592 candidate-only records, 1,443 records with no candidate, and 8 unavailable responses or CoT traces are excluded. Candidate-only annotations remain in the source data.

The 95\% interval uses a paired item-level percentile bootstrap of $g_i$
with 10,000 resamples and seed 20260921. In the archived annotations, forty-two discrepancies between
declared best states and recorded exploration paths were audited using
the complete-state maximum rule, with the event labels preserved. The
retained checkpoint may precede later reasoning or the final answer.
Consequently, this gap describes quality preservation at an observable
Aha checkpoint. Intervals quantify sampling variability in those scorable
records and retain the historical scoring-anchor and annotation-boundary
differences across corpora.

\subsection{Raw Aha Counts and A/E/V Changes}
\label{app:process-counts}

\begin{table*}[t]
\centering
\small
\setlength{\tabcolsep}{4pt}
\renewcommand{\arraystretch}{1.08}

\begin{tabular*}{\textwidth}{@{\extracolsep{\fill}}lrrrrrrrr@{}}
\toprule
Model & C & S & B & Functional & Missed & Unresolved & Ineffective & Regressive \\
\midrule
GPT-5.5 & 141 & 79 & 39 & 36 & 3 & 4 & 23 & 13 \\
GPT-5.6 Sol & 116 & 78 & 37 & 33 & 4 & 6 & 26 & 9 \\
Gemini 2.5 Flash & 1898 & 494 & 362 & 286 & 76 & 24 & 80 & 28 \\
Gemini 3.5 Flash & 843 & 291 & 172 & 148 & 24 & 15 & 74 & 30 \\
Claude Opus 4.8 & 859 & 447 & 238 & 197 & 41 & 32 & 113 & 64 \\
Kimi K2.6 & 2624 & 385 & 262 & 150 & 112 & 36 & 52 & 35 \\
Kimi K3 & 4266 & 1156 & 593 & 398 & 195 & 359 & 159 & 45 \\
Qwen 3.7 Plus & 2427 & 845 & 486 & 298 & 188 & 191 & 137 & 31 \\
Qwen 3.8 Max & 864 & 383 & 229 & 173 & 56 & 94 & 56 & 4 \\
MiniMax M3 & 1530 & 658 & 346 & 263 & 83 & 149 & 134 & 29 \\
Doubao 2.1 Pro & 3597 & 1570 & 1084 & 431 & 653 & 163 & 265 & 58 \\
\bottomrule
\end{tabular*}
\caption{Raw Aha event counts. C: Aha candidates; S: substantive events; B: better-state discoveries. The five outcome columns sum to S.}
\label{tab:process-raw}
\par\smallskip
\begin{minipage}{\textwidth}\footnotesize\raggedright
Candidate-only records contribute to C and remain outside S.
\end{minipage}
\end{table*}

Table~\ref{tab:process-raw} provides the counts underlying the eleven-model
process comparison; Table~\ref{tab:aev-full} gives the corresponding mean
component changes over substantive events. Event counts and character
densities weight different features of the recorded traces. For example,
Opus has 197 Functional events over 123,984 CoT characters, while Gemini 3.5 Flash has 148 over 596,859 characters. The resulting FED values are 15.889 and 2.480 per 10,000 characters, respectively.

\begin{table*}[t]
\centering
\small
\setlength{\tabcolsep}{4pt}
\renewcommand{\arraystretch}{1.08}

\begin{tabular*}{\textwidth}{@{\extracolsep{\fill}}lrrrr@{}}
\toprule
Model & $\Delta q_A$ & $\Delta q_E$ & $\Delta q_V$ & $\Delta Q_{\mathrm{total}}$ \\
\midrule
GPT-5.5 & 0.0886 & 0.0190 & 0.1646 & 0.0907 \\
GPT-5.6 Sol & 0.0897 & 0.0224 & 0.1987 & 0.1036 \\
Gemini 2.5 Flash & 0.1215 & 0.0805 & 0.3391 & 0.1803 \\
Gemini 3.5 Flash & 0.2131 & 0.1289 & 0.2251 & 0.1890 \\
Claude Opus 4.8 & 0.1208 & 0.0805 & 0.1611 & 0.1208 \\
Kimi K2.6 & 0.2571 & 0.1636 & 0.2604 & 0.2271 \\
Kimi K3 & 0.1635 & 0.0679 & 0.1254 & 0.1189 \\
Qwen 3.7 Plus & 0.2391 & 0.1612 & 0.2021 & 0.2008 \\
Qwen 3.8 Max & 0.2037 & 0.1534 & 0.1234 & 0.1601 \\
MiniMax M3 & 0.1125 & 0.0315 & 0.2523 & 0.1321 \\
Doubao 2.1 Pro & 0.3962 & 0.1158 & 0.3519 & 0.2880 \\
\bottomrule
\end{tabular*}
\caption{Mean A/E/V changes between the comparison state used for event classification, $\widetilde H_k$, and the pre-event state, $H_k$.}
\label{tab:aev-full}
\end{table*}

\begin{figure}[!htbp]
    \centering
    \includegraphics[width=\columnwidth]{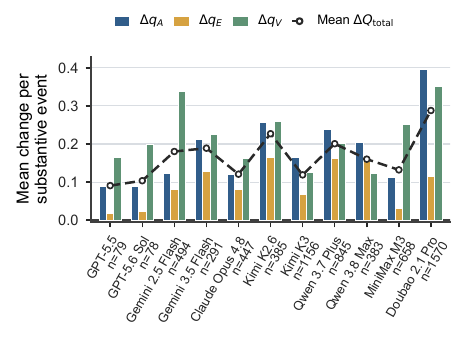}
    \caption{A/E/V decomposition of the quality differences used for event classification. Full A/E/V values are provided in Table~\ref{tab:aev-full}.}
    \label{fig:aev-decomposition}
\end{figure}

\paragraph{Aha moments involve changes beyond the final answer.}
The A/E/V decomposition in Figure~\ref{fig:aev-decomposition}
uses the comparison states for event classification and shows that
revisions during Aha moments extend beyond changes in
the candidate answer. Across models, improvements in reasoning-state
quality may arise primarily from answer quality, evidence coverage, or
candidate--evidence consistency. Some models show larger gains in the
consistency between their hypotheses and the available evidence, whereas
others derive more of their improvement from changes in the answer state.

This distinction explains why answer-level self-correction alone provides
an incomplete account of revision behavior. During an Aha moment,
a model may reinterpret a key clue, eliminate a candidate that conflicts
with the evidence, or construct a more coherent explanation without
immediately reaching the correct final answer. Conversely, a change in
the candidate answer does not necessarily imply an overall improvement
in the reasoning state. Jointly evaluating answer quality, evidence use,
and candidate--evidence consistency therefore distinguishes superficial
answer switching from evidence-grounded substantive revision.

\subsection{A/E/V Ablation and Weight Sensitivity}
\label{app:aev-ablation}

We evaluate the contribution of the three state-quality components using
the same 110 reasoning traces and 62 human-confirmed effective revision
events. The table reports event-level matching results; the Wait-only
baseline is omitted from this comparison.

\begin{table*}[t]
\centering
\small
\setlength{\tabcolsep}{5pt}
\renewcommand{\arraystretch}{1.08}

\begin{tabular*}{\textwidth}{@{\extracolsep{\fill}}lrrrrrrr@{}}
\toprule
Method & TP & FP & FN & Predicted & Precision & Recall & F1 \\
\midrule
Full A/E/V & 49 & 11 & 13 & 60 & 81.7 & 79.0 & 80.3 \\
w/o A & 45 & 14 & 17 & 59 & 76.3 & 72.6 & 74.4 \\
w/o E & 36 & 17 & 26 & 53 & 67.9 & 58.1 & 62.6 \\
w/o V & 34 & 10 & 28 & 44 & 77.3 & 54.8 & 64.2 \\
Simple reflection cues & 48 & 37 & 14 & 85 & 56.5 & 77.4 & 65.3 \\
\bottomrule
\end{tabular*}
\caption{Human-evaluated ablation of the A/E/V criteria. TP, FP, and FN
denote true positives, false positives, and false negatives. Precision is
$\mathrm{TP}/(\mathrm{TP}+\mathrm{FP})$, recall is $\mathrm{TP}/62$, and
F1 is their harmonic mean.}
\label{tab:aev-ablation}
\par\smallskip
\begin{minipage}{\textwidth}\footnotesize\raggedright
All rates are percentages. The table uses the human-reviewed event reference
for the 110-trace diagnostic sample; each reference event is matched at most
once. Wait-only matching is excluded.
\end{minipage}
\end{table*}

\subsubsection{Sensitivity to A/E/V Weights}
\label{app:aev-weight-sensitivity}

We assessed weighting sensitivity on a fixed subset of 220 CoT traces
from eleven models (20 per model; 344 substantive events), sampled
across languages, trace lengths, and event-count strata. Event boundaries,
candidate and substantive judgments, adoption decisions, and annotated
comparison states were held fixed. We compared equal weights with
three settings emphasizing $A$, $E$, or $V$, and 1,000 weight vectors
drawn uniformly from the probability simplex. For each setting, we
recomputed event outcomes, BDR, DAR, FED, and NQGD.
Outcome labels use the comparison states in Section~\ref{sec:diagnostic-metrics};
NQGD always compares actual post-event and pre-event states.

\begin{table*}[t]
\centering
\small

\begin{tabular}{lccrrrr}
\toprule
Setting & Weights $(w_A,w_E,w_V)$ & Reclassified (\%) & BDR & DAR & FED & NQGD \\
\midrule
Equal weights & $(1/3,1/3,1/3)$ & 0.00 & 55.33 & 69.51 & 4.831 & 0.902 \\
$A$ emphasis & $(0.6,0.2,0.2)$ & 0.00 & 55.33 & 69.51 & 4.831 & 0.998 \\
$E$ emphasis & $(0.2,0.6,0.2)$ & 5.88 & 52.24 & 68.12 & 4.104 & 0.547 \\
$V$ emphasis & $(0.2,0.2,0.6)$ & 1.48 & 56.55 & 69.43 & 4.916 & 1.160 \\
\bottomrule
\end{tabular}
\caption{Sensitivity to A/E/V weights on the fixed subset.
Entries are averaged equally across the eleven models.
Reclassified denotes the percentage of substantive events whose outcome
label changes relative to equal weighting. BDR and DAR are percentages;
FED and NQGD are per 10,000 CoT characters.}
\label{tab:aev-weight-sensitivity}
\end{table*}

Table~\ref{tab:aev-weight-sensitivity} shows limited changes in event
classification and discovery/adoption diagnostics. Across the three fixed
alternatives, mean event reclassification ranged from 0 to 5.88\%,
while mean BDR and DAR changed by at most 3.08 and 1.39 percentage
points, respectively. Across the 1,000 sampled weight vectors, the mean
reclassification rate was 3.04\%, with a maximum of 6.53\%.
The minimum model-ranking Spearman correlations relative to equal
weighting were 0.909 for BDR, 0.973 for DAR, and 0.991 for FED.
NQGD rankings were more sensitive to the weights ($\rho=-0.300$--$1.000$),
with the lowest correlations occurring under strongly $E$-dominant weights.
These results describe the fixed sampled subset.

\subsection{History and Candidate Selection}
\label{app:history-intervention}

\paragraph{Setup.}
We retain the two Qwen models fixed in the development protocol for
mechanism replication within one model family. Each of the 400 puzzles is
screened for the first explicit conflict following a preferred complete
answer; the checkpoint ends before repair. Boundaries are located and
reviewed with GPT-5.5 assistance. Analyses use complete four-condition
pairs: 60/22 incorrect/correct checkpoints for Q3.7 and 44/31 for Q3.8.
Search uses temperature 0.7 and an 8,192-token cap. SB deduplicates three
independent rollouts, randomizes candidate labels, and uses a fresh
same-model selector (temperature 0, 4,096-token cap).
All conditions use OpenRouter/Alibaba with fixed settings within model:
\texttt{qwen/qwen3.7-plus} disables provider thinking, whereas
\texttt{qwen/qwen3.8-max-0902} requires \texttt{minimal} reasoning,
stored separately from CoT. These exploratory replays use frozen
checkpoints; equivalence to the original API snapshots is unverified.

\paragraph{Metrics and tests.}
Table~\ref{tab:history-full} reports all four conditions. Retention is an
exact normalized match to the checkpoint candidate; AIR measures repair
of incorrect candidates, and ADR measures errors on correct controls.
Discovery and adoption are measured on each model's generated pools.
S1 retention comparisons use two-sided exact McNemar tests with Holm
correction across the two models; secondary tests are unadjusted.
AIR gains have $p=.125$ and $.625$ for Q3.7/Q3.8.
P0 jointly changes history content and length; among accepted outputs,
SB uses 3.57--4.74 times the S1 tokens.

\begin{table*}[!t]
\centering\small
\setlength{\tabcolsep}{4pt}

\begin{tabular*}{\textwidth}{@{\extracolsep{\fill}}llrrrrr@{}}
\toprule
Model & Condition & Ret.$\downarrow$ & AIR$\uparrow$ & ADR$\downarrow$ & Found & Adopted \\
\midrule
Q3.7 & P1-S1 & 34/60 & 6/60 & 1/22 & -- & -- \\
Q3.7 & P0-S1 & 15/60 & 11/60 & 4/22 & -- & -- \\
Q3.7 & P1-SB & 29/60 & 7/60 & 1/22 & 8/60 & 7/8 \\
Q3.7 & P0-SB & 16/60 & 10/60 & 3/22 & 16/60 & 10/16 \\
\midrule
Q3.8 & P1-S1 & 21/44 & 5/44 & 4/31 & -- & -- \\
Q3.8 & P0-S1 & 4/44 & 7/44 & 9/31 & -- & -- \\
Q3.8 & P1-SB & 22/44 & 9/44 & 5/31 & 9/44 & 9/9 \\
Q3.8 & P0-SB & 10/44 & 9/44 & 9/31 & 9/44 & 9/9 \\
\bottomrule
\end{tabular*}
\caption{Complete paired intervention results. Entries are numerator/denominator
counts. Ret.\ and AIR use incorrect checkpoints; ADR uses correct controls.
Found counts gold-containing pools; Adopted conditions on those pools ($K=3$).
All four conditions use the same complete items within each model.}
\label{tab:history-full}
\end{table*}

\subsection{Paired Information Assistance and Event Denominators}
\label{app:information-details}
For each contrast in Table~\ref{tab:information-paired}, the first state is
the earlier condition and the second is the later condition. Thus $WC$
counts gains and $CW$ counts losses. Accuracy changes and their intervals
are in percentage points. Intervals use 10,000 paired item bootstrap
resamples with seed 20260920. Exact two-sided McNemar $p$ values are
unadjusted across the displayed comparisons. Bootstrap intervals and
discrete exact tests use different inferential constructions.

\begin{table*}[!t]
\centering
\small
\setlength{\tabcolsep}{4pt}
\renewcommand{\arraystretch}{1.08}

\begin{tabular*}{\textwidth}{@{\extracolsep{\fill}}llrrrrrcr@{}}
\toprule
Model & Contrast & CC & CW & WC & WW & $\Delta$ & 95\% CI & $p$ \\
\midrule
Qwen 3.7 & T1-T0 & 41 & 6 & 20 & 33 & +14 & [4, 24] & 0.0094 \\
Qwen 3.7 & T2-T1 & 56 & 5 & 10 & 29 & +5 & [-2, 13] & 0.3018 \\
Qwen 3.7 & T2-T0 & 40 & 7 & 26 & 27 & +19 & [8, 30] & 0.0013 \\
Qwen 3.8 & T1-T0 & 56 & 7 & 17 & 20 & +10 & [1, 19] & 0.0639 \\
Qwen 3.8 & T2-T1 & 69 & 4 & 5 & 22 & +1 & [-5, 7] & 1.0000 \\
Qwen 3.8 & T2-T0 & 59 & 4 & 15 & 22 & +11 & [3, 19] & 0.0192 \\
Opus 4.8 & T1-T0 & 35 & 4 & 20 & 41 & +16 & [7, 25] & 0.0015 \\
Opus 4.8 & T2-T1 & 45 & 10 & 8 & 37 & -2 & [-10, 6] & 0.8145 \\
Opus 4.8 & T2-T0 & 32 & 7 & 21 & 40 & +14 & [4, 24] & 0.0125 \\
\bottomrule
\end{tabular*}
\caption{Paired information-assistance results, with 100 paired puzzles in every row.}
\label{tab:information-paired}
\end{table*}
\begin{table*}[!t]
\centering
\small
\setlength{\tabcolsep}{4pt}
\renewcommand{\arraystretch}{1.08}

\begin{tabular*}{\textwidth}{@{\extracolsep{\fill}}llrrrrr@{}}
\toprule
Model & Input & S & B & F & BDR & DAR \\
\midrule
Qwen 3.7 Plus & T0 & 310 & 156 & 90 & 50.3 & 57.7 \\
Qwen 3.7 Plus & T1 & 97 & 79 & 59 & 81.4 & 74.7 \\
Qwen 3.7 Plus & T2 & 111 & 82 & 66 & 73.9 & 80.5 \\
Qwen 3.8 Max & T0 & 182 & 111 & 85 & 61.0 & 76.6 \\
Qwen 3.8 Max & T1 & 87 & 66 & 52 & 75.9 & 78.8 \\
Qwen 3.8 Max & T2 & 80 & 65 & 51 & 81.3 & 78.5 \\
Opus 4.8 & T0 & 99 & 55 & 48 & 55.6 & 87.3 \\
Opus 4.8 & T1 & 30 & 27 & 24 & 90.0 & 88.9 \\
Opus 4.8 & T2 & 26 & 18 & 15 & 69.2 & 83.3 \\
\bottomrule
\end{tabular*}
\caption{Event denominators for information assistance. BDR is $100B/S$; DAR is $100F/B$.}
\label{tab:information-events}
\par\smallskip
\begin{minipage}{\textwidth}\footnotesize\raggedright
T0 uses historical annotations. Opus T1 and T2 retain 13/30 and 7/26 boundary-uncertain substantive events. Supplied facts and mechanisms are not spontaneous Aha discoveries.
\end{minipage}
\end{table*}

All nine process rows are retained as descriptive measurements with their
annotation provenance. For Opus T2, excluding the seven boundary-uncertain
events changes BDR from 69.2\% to 57.9\% and DAR from 83.3\% to
90.9\%. This sensitivity leaves 19 substantive events and illustrates
the influence of event boundaries on the small denominators. The Opus
T1 pilot has 13 uncertain events among 30. Cross-condition process
differences retain annotation-source and scoring-anchor differences.
The paired accuracy results show consistent gains after observation facts
on these puzzles, followed by smaller, heterogeneous changes after the
mechanism cue: $+5$, $+1$, and $-2$ points for the three models.

\clearpage
\makeatletter
\global\@colht\textheight
\global\@colroom\textheight
\global\vsize\textheight
\makeatother
\section{Extended Related Work}
\label{app:extended-related-work}

\subsection{Rebus and Multimodal Puzzle Benchmarks}

Rebus puzzles require models to jointly reason over visual cues, textual structure, spatial relations, and linguistic associations, and have therefore been increasingly used to evaluate abstract and lateral reasoning in multimodal models. Existing benchmarks approach this problem from different perspectives. REBUS and COLUMBUS primarily evaluate puzzle-solving ability through final-answer correctness, while Puzzled by Puzzles extends the evaluation to a broader range of visual metaphors and linguistic wordplay. In the Chinese setting, the Pun Rebus Art Dataset further emphasizes homophony, cultural symbols, and implied meanings
\cite{gritsevskiy2024rebus,kraaijveld2025columbus,lee2025puzzled,zhang2025chinese}.
Re-Bus extends coverage across puzzle styles and domains and explores description- and program-based reasoning, while RebusBench examines the integration of visual cues with linguistic knowledge \cite{das2025rebus,kasaei2026rebusbench}. Puzzled by Puzzles also evaluates skill-guided prompting and iterative refinement \cite{lee2025puzzled}.
More recently, Reasoning Riddles separates answer correctness from explanation quality and compares reasoning processes elicited by different prompting strategies
\cite{movva2025reasoning}.

Despite this progress, most existing work evaluates either the final answer or the overall quality of a completed explanation, with limited attention to how hypotheses evolve within a single reasoning trajectory. AhaPro adds a process-diagnostic layer for CoT traces by separating candidate reconsideration from substantive Aha, tracing A/E/V states and the discovery--adoption gap, and testing history and candidate-search interventions.

\subsection{Aha Moments and Chain-of-Thought}

In human insight research, rebus puzzles have been studied as a representative class of representational-restructuring problems, where successful solutions often require breaking away from an initially dominant interpretation of textual or visual elements
\cite{macgregor2008rebus}.
With the emergence of reasoning models that produce extended visible reasoning traces, mid-trace behaviors such as ``wait'' and explicit reconsideration have increasingly been discussed as Aha moments
\cite{deepseek2025r1}.
Recent work analyzes Aha-related linguistic patterns, uncertainty, reasoning collapse, and latent representations \cite{yang2025ahamoments}.
Recent studies, however, suggest that such reasoning shifts do not necessarily improve task performance, and that apparently reflective or self-verifying steps may sometimes be merely decorative rather than functionally consequential
\cite{daliberti2026illusion,zhao2025fakeaha}.

Chain-of-Thought (CoT) provides an observable interface for studying these behaviors. CoT prompting, including simple instructions such as ``Let's think step by step,'' can elicit explicit intermediate reasoning, making hypothesis formation, verification, rejection, and revision observable over time
\cite{wei2022cot,kojima2022zeroshot}.
Similar effects have also been reported in rebus solving, where prompting strategies can substantially alter both the generated reasoning process and task performance
\cite{movva2025reasoning}.
Accordingly, AhaPro uses prompted visible CoT as the behavioral trace for identifying and evaluating state transitions during Aha moments. Importantly, these traces are treated as observable outputs rather than faithful readouts of the model's internal computation
\cite{turpin2023unfaithful}.

ROSCOE evaluates semantic consistency, logicality, and informativeness in step-by-step reasoning \cite{golovneva2023roscoe}.
Other process evaluations assess step correctness and informativeness \cite{prasad2023receval}, validity and redundancy \cite{xia2025reasoneval}, and error localization \cite{zheng2025processbench}.
PRMBench evaluates process reward models across simplicity, soundness, and sensitivity \cite{song2025prmbench}.
Process supervision uses feedback on intermediate reasoning steps to train reward models for mathematical reasoning \cite{lightman2024verify}.
Studies of long reasoning models also identify redundant solutions with limited gains in accuracy or strategy diversity \cite{chen2025overthinking}.

\subsection{Self-Correction, Search, and Reasoning Interventions}

Prior work has shown that intrinsic self-correction is unreliable in the absence of external feedback and may even cause models to revise initially correct answers
\cite{huang2024selfcorrect}.Providing error locations can improve subsequent correction on reasoning tasks \cite{tyen2024errorlocation}.
Self-Refine uses model-generated feedback for iterative revision \cite{madaan2023selfrefine}, while SCoRe trains self-correction through multi-turn reinforcement learning \cite{kumar2025score}.
At the same time, approaches such as Tree of Thoughts demonstrate that branching over multiple candidate paths, evaluating alternatives, and backtracking can alleviate some limitations of single-path autoregressive reasoning
\cite{yao2023tot}.
Studies of verifier-guided search in mathematical reasoning identify candidate misranking and erroneous pruning of valid paths as factors limiting the benefits of increased search \cite{yu2025scalingflaws}.
Together, these findings suggest that a failure to correct after detecting a conflict may arise at different stages: the model may fail to generate a better hypothesis, or it may discover a better alternative but fail to adopt it.

Controlled interventions on CoT provide a way to distinguish these possibilities. Previous studies have modified, perturbed, or truncated intermediate reasoning and examined subsequent changes in model predictions to test whether visible reasoning steps are functionally used
\cite{lanham2023faithfulness,zhao2025fakeaha}.
Recent work on Aha moments has similarly compared spontaneous reasoning shifts with externally induced reconsideration
\cite{daliberti2026illusion}.
Building on these approaches, AhaPro constructs paired interventions from a shared conflict checkpoint and manipulates two factors: whether the preceding reasoning prefix is retained, and whether subsequent reasoning is supported by independent candidate search. This design helps distinguish failures associated with path dependence, restricted search, and the inability to adopt an already discovered better hypothesis.

\end{document}